\pdfoutput=1
\documentclass[11pt]{article}

\usepackage{EMNLP2022}

\usepackage{times}
\usepackage{latexsym}

\usepackage[T1]{fontenc}
\usepackage[utf8]{inputenc}

\usepackage{microtype}

\usepackage{inconsolata}

\usepackage[utf8]{inputenc}

\usepackage{microtype}
\usepackage{rotating}
\usepackage{blindtext}
\usepackage{amsmath}
\usepackage{amsfonts}
\usepackage{graphicx}
\usepackage{mathtools}
\usepackage{booktabs}
\usepackage{relsize}
\usepackage{float}
\usepackage{graphics}
\usepackage{multirow}
\usepackage{algorithm}
\usepackage{algpseudocode}
\usepackage[utf8]{inputenc}

\usepackage{threeparttable}
\usepackage{booktabs, caption, makecell}

\usepackage[title]{appendix}
\usepackage[cal=cm]{mathalfa}
\usepackage[colorinlistoftodos]{todonotes}
\usepackage{multirow}
\usepackage[export]{adjustbox}
\usepackage{subcaption}
\usepackage{pifont}
\usepackage{tabularx}
\usepackage{longtable}
\usepackage{tikz}
\usepackage{enumitem}
\usepackage{diagbox}

\usepackage{colortbl,dcolumn}

\newcommand{\ignore}[1]{}

\usepackage{microtype}
\usepackage[subtle]{savetrees}
\title{What Do Compressed Multilingual Machine Translation Models Forget?}

\author{Alireza Mohammadshahi\thanks{~~~Work done during an internship at NAVER LABS Europe.}~~$^{1,2,3}$ ~~Vassilina Nikoulina$^1$ ~~Alexandre Berard$^1$ \\ \textbf{Caroline Brun}$^1$ ~~\textbf{James Henderson}$^2$ ~~\textbf{Laurent Besacier}$^1$ \vspace{0.1cm}\\ 
 $^1$ NAVER LABS Europe ~~~~~ $^2$ IDIAP Research Institute ~~~~~ $^3$ EPFL\\
 \texttt{\{first.last\}@naverlabs.com}\\\texttt{\{alireza.mohammadshahi,james.henderson\}@idiap.ch}
}

\begin{document}
\maketitle

\begin{abstract}

Recently, very large pre-trained models achieve state-of-the-art results in various natural language processing~(NLP) tasks, but their size makes it more challenging to apply them in resource-constrained environments. Compression techniques allow to drastically reduce the size of the models and therefore their inference time with negligible impact on top-tier metrics.  However,  the general performance averaged across multiple tasks and/or languages may hide a drastic performance drop on under-represented features, which could result in the amplification of biases encoded by the models. 
In this work, we assess the impact of compression methods on Multilingual Neural Machine Translation models~(MNMT) for various language groups, gender, and semantic biases by extensive analysis of compressed models on different machine translation benchmarks,  i.e. FLORES-101, MT-Gender, and DiBiMT.  We show that the performance of under-represented languages drops significantly, while the average BLEU metric only slightly decreases. Interestingly, the removal of noisy memorization with compression leads to a significant improvement for some medium-resource languages. Finally, we demonstrate that compression amplifies intrinsic gender and semantic biases, even in high-resource languages.\footnote{We release our implementation at \url{https://github.com/alirezamshi/bias-compressedMT}.}

\end{abstract}

\section{Introduction}

Over the recent years, pre-trained Transformer~\cite{transformer} models have reached a substantial improvement in a variety of Natural Language Processing~(NLP) tasks. This improvement mostly comes from increasing their parameter size~\cite{devlin-etal-2019-bert,m2m-100,NEURIPS2020_1457c0d6,https://doi.org/10.48550/arxiv.2205.01068}  which escalates the cost of training~\cite{NEURIPS2019_dc6a7e65,strubell-etal-2019-energy,carbontrain}, and hurts the memory footprint and latency at inference ~\cite{dai-etal-2019-transformer,m2m-100,deepnet}. Specially in Neural Machine Translation~(NMT) task, massively MNMT models~\cite{aharoni-etal-2019-massively,m2m-100,mbart,zhang-etal-2020-improving} demonstrated promising results. They have been shown particularly interesting for low-resource languages which benefit a lot from knowledge transfer. On the other hand, it has also been observed that the \textit{curse of multilinguality} may hurt the performance in high-resource languages. The strategy employed to overcome this problem~\cite{aharoni-etal-2019-massively,m2m-100, goyal-etal-2021-large} is to scale up the number of parameters, thus attaining state-of-the-art performance in both high and low-resource languages.

Consequently, efficient inference with these very large models has become a crucial problem. This challenge can be overcome through model compression, e.g. knowledge distillation~\cite{kim-rush-2016-sequence,distilbert,https://doi.org/10.48550/arxiv.2012.13866,wang-etal-2021-selective}, pruning~\cite{h.2018to,frankle2018the,behnke-heafield-2020-losing,zhang-etal-2021-enlivening}, and quantization~\cite{https://doi.org/10.48550/arxiv.1802.00150,https://doi.org/10.48550/arxiv.2004.09602,bondarenko-etal-2021-understanding,pmlr-v139-kim21d,https://doi.org/10.48550/arxiv.2203.10705,https://doi.org/10.48550/arxiv.2203.15996,yao2022zeroquant}. These methods can be applied with a little loss in top-line metrics, while reducing the memory-footprint, and enhancing inference time. 
However, recent work~\cite{https://doi.org/10.48550/arxiv.2010.03058,ahia-etal-2021-low-resource,xu-etal-2021-beyond,https://doi.org/10.48550/arxiv.2110.08419,renduchintala-etal-2021-gender} has demonstrated that under-represented features can suffer from a drastic decrease in performance which is not necessarily reflected by global (aggregated) metrics.
In multilingual NMT, the overall metrics are often reported as an average across all the language pairs, where the performance between individual language pairs can vary a lot. Therefore it is even more critical to understand what would be the exact impact of compression on multilingual NMT models, beyond the aggregated metrics.

In this work, we illustrate the impacts of applying compression methods to massively multilingual NMT models, that are pre-trained in a great number of languages in several domains. To the best of our knowledge, this is the first attempt to analyze how compression impacts massively multilingual models. We hope it could be a starting point to bringing a comprehensive understanding between fairness and compression in multilingual NMT models. In this study, we concentrate on  \textit{light} compression techniques, specifically post-training quantization and magnitude pruning without any further fine-tuning.\footnote{The reason is that fine-tuning MNMT models is extremely computationally demanding.} We exploit the recent and largest MNMT model, M2M-100~\cite{m2m-100} that covers 100 languages and contains nearly 12B parameters and analyze the impact of compression on different language pairs evaluated on FLORES-101 benchmark ~\cite{flores101} (covering 101 languages). We also consider MT-Gender~\cite{stanovsky-etal-2019-evaluating} and DiBiMT~\cite{dibimt} benchmarks allowing us to assess different types of biases that could be present in the data and MNMT model. 
To sum up, our contributions are as follows:

\begin{itemize}[noitemsep,topsep=0pt,parsep=0pt,partopsep=0pt]
    \item We conduct extensive analysis on the effects of \textit{light} compression methods for massively multilingual NMT models.
    \item On FLORES-101~\cite{flores101}, we discover that while the overall performance is barely impacted by the compression, a subset of language pairs corresponding to under-represented languages during training suffers an extreme drop in performance.
    \item Also, we observe an important improvement for some language pairs after the compression. We hypothesize that this is due to the removal of noisy memorization.
    \item We show that the compression amplifies gender and semantic biases, hidden in MNMT models across several high-resource languages by evaluating on MT-Gender, and DiBiMT benchmarks.
\end{itemize}

In section~\ref{sec:method}, we describe \textit{light} compression methods we rely on, and MNMT model. Section~\ref{sec:setup} presents our experimental setup and evaluation benchmarks. Section~\ref{sec:result} shows the analysis of the impact of the compression for NMT benchmarks. 

\section{Model and Compression Techniques}
\label{sec:method}

\subsection{M2M-100 Model}

We assume that potential biases, discovered after the compression are mostly related to the training data, than the model architecture, as previous work~\cite{https://doi.org/10.48550/arxiv.2010.03058} demonstrated for the image classification task. \\
So, we use M2M-100~\cite{m2m-100}, as it is the best performing and the largest massively multilingual MT model, which covers more than 10K language directions, including a great number of low and medium-resource language pairs. Other previous work~\cite{aharoni-etal-2019-massively,mbart} cover fewer languages, especially from low and medium-resource languages, and have worse results compared to M2M-100. \\
M2M-100 is trained on large-scale multilingual corpora~\cite{el-kishky-etal-2020-ccaligned,schwenk-etal-2021-ccmatrix} with a novel data mining procedure, that uses language similarities. The biggest model introduced consists of 24 encoder, and 24 decoder Transformer~\cite{transformer} layers. Using several scaling techniques, it is trained with nearly 12B parameters. We refer to \citet{m2m-100} for more details. In all our experiments, we exploit the largest M2M-100 model.

\subsection{Light Compression Techniques}

Compression techniques without any further fine-tuning are defined as \textit{light} compression methods. We do not fine-tune the compressed models due to the massive computation cost, as we have to fine-tune the model for all language pairs to provide a fair comparison.~\footnote{Additionally, the exact and original training data is required to alleviate the additional bias added by fine-tuning, but M2M-100 authors do not provide the exact data e.g. back-translation.} We discuss our methods in the following paragraphs.

\textbf{Magnitude Pruning}
 is a popular technique for both memory footprint reduction and inference speed-up. It reduces the model size by removing redundant nodes that do not contribute to the resulting performance. It usually achieves comparable results with state-of-the-art models with further fine-tuning~\cite{h.2018to,https://doi.org/10.48550/arxiv.1902.09574,https://doi.org/10.48550/arxiv.2106.08962,ahia-etal-2021-low-resource}. In this work, we apply post-training magnitude pruning for each layer of Transformer (including Embedding layers). Given $\Theta_l$ as the parameters of Transformer layer $l$ and $p$ as the sparsity ratio, the output of the pruning function is $\Theta'_l$ where $p$ percentage of weights sets to zero.\footnote{Preliminary experiments showed that pruning based on Transformer layer results in a better performance than other alternatives e.g. separate pruning of self-attention and feed-forward layers. The comparison is provided in Appendix~\ref{app:magn-strategy}.}

\begin{figure}
\centering
\begin{subfigure}[b]{\linewidth}
    \centering
    \includegraphics[width=0.8\linewidth]{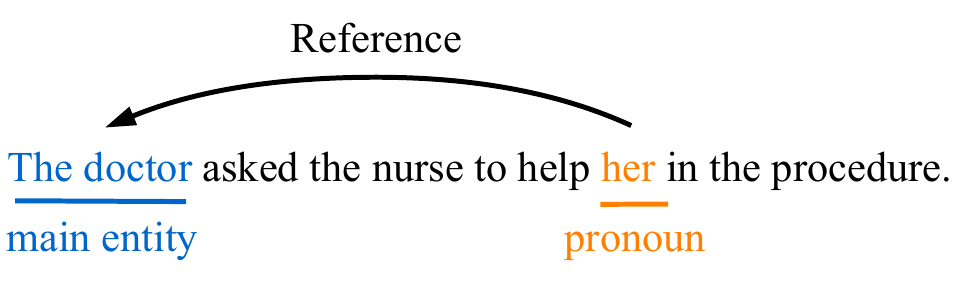}
    \caption{MT-Gender example: for a correct translation, system will have to link English pronoun 'her' to 'doctor'.}
    \label{fig:ex:mt-gender}
\end{subfigure}\vspace{0.2cm}
\begin{subfigure}[b]{\linewidth}
    \centering
    \includegraphics[width=0.5\linewidth]{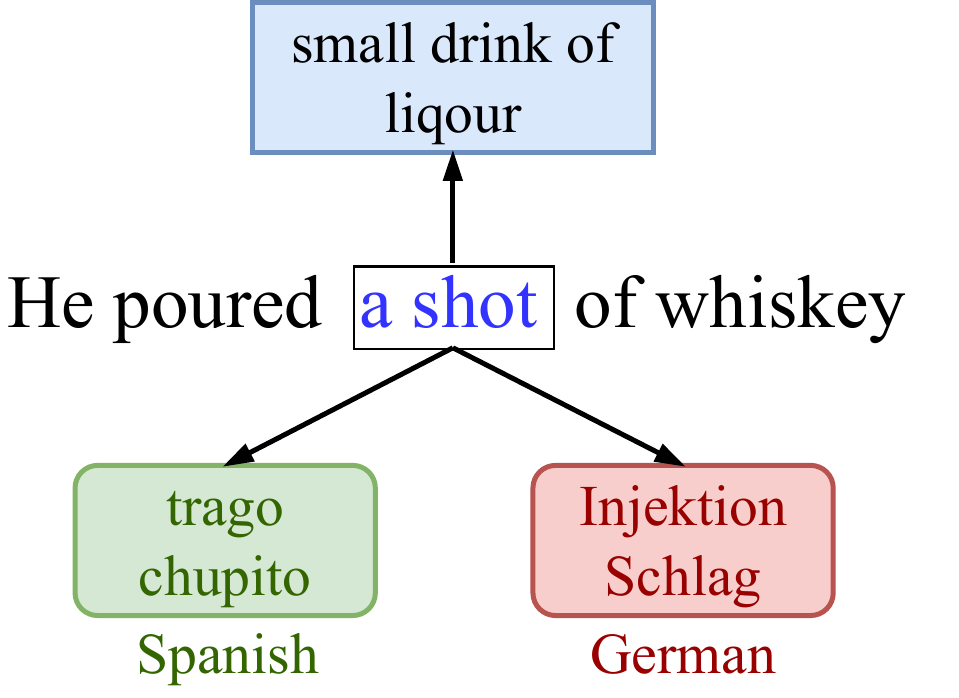}
    \caption{DiBiMT Example. German instance contains wrong word senses, while Spanish one is correct.}
    \label{fig:ex:dibimt}
\end{subfigure}
\caption{Samples of MT-Gender~\cite{stanovsky-etal-2019-evaluating} and DiBiMT~\cite{dibimt} benchmarks.}
\label{fig:example-data}
\end{figure}

\textbf{Post-Training Quantization}
Recent work applies post-training, and training-aware quantization to pre-trained machine translation and language models~\cite{https://doi.org/10.48550/arxiv.2004.09602,https://doi.org/10.48550/arxiv.2106.08962,https://doi.org/10.48550/arxiv.2101.09671,bondarenko-etal-2021-understanding,https://doi.org/10.48550/arxiv.2203.05740}, and achieves promising results while lowering the inference latency, and the model size.
In this work, we exploit the post-training quantization method proposed by \citet{https://doi.org/10.48550/arxiv.2004.09602}, converting all weights and activations from 32-bit floating-point values to an 8-bit fixed-point integer. Specifically, it quantizes linear layers input and weights, matrix multiplications, and the residual summations for Transformer~\cite{transformer}.
\section{Experimental Setup}
\label{sec:setup}

\subsection{Evaluation Benchmarks}

We analyze our compressed models on three different NMT benchmarks. We exploit FLORES-101~\cite{flores101} to study the model behavior based on the amount of available resources for each language. MT-Gender~\cite{stanovsky-etal-2019-evaluating} is used to study the impact of compression on gender bias. Finally, we evaluate on DiBiMT~\cite{dibimt} to illustrate the compression effect on semantic biases.

\paragraph{FLORES-101} is a many-to-many NMT evaluation benchmark, including sentences extracted from English Wikipedia. It is translated into 101 languages by human translators, enabling 10,100 language directions to be evaluated. In this paper, we evaluate our models on {\tt devtest} subset of the FLORES-101~\cite{flores101} benchmark. This benchmark provides test sets comparable across all the language pairs, and thus allows us to assess to what extent each language pair gets impacted by the compression techniques. 

\paragraph{MT-Gender} 

~\cite{stanovsky-etal-2019-evaluating} is an English-centric multilingual NMT benchmark for evaluating gender bias in multiple target languages: Arabic, Ukrainian, Hebrew, Russian, Italian, French, Spanish, and German. The method relies on automatic alignment and morphological analysis, without the need for gold translations.\footnote{For each instance, the main entity is attached to a pronoun, and the side entity attempts to distort the translation. With the use of automatic alignment and morphological analysis, the translated gender is extracted.} An example is shown in Figure~\ref{fig:ex:mt-gender}. Later, \citet{kocmi-etal-2020-gender} extends the benchmark by adding Czech and Polish languages. We choose MT-Gender as it covers more languages compared to other existing MT gender bias benchmarks~\cite{bentivogli-etal-2020-gender,renduchintala-etal-2021-gender,savoldi-etal-2022-morphosyntactic}.

\paragraph{DiBiMT} is the first fully manually-crafted NMT benchmark for evaluating word sense disambiguation on five high-resource languages: Chinese, German, Italian, Russian, and Spanish~\cite{dibimt}, where the source language is English. Besides, they propose several bias evaluation metrics to compare different models~(defined in Section~\ref{sec:dibimt}). As shown in Figure~\ref{fig:ex:dibimt}, given English source sentence, specific word~($w_i$) with associated synset~($\sigma$), and language $L$, set of {\tt GOOD}, and {\tt BAD} translation candidates include sentences that do and do not contain set of correct translation of $\sigma$ in language $L$, respectively. More details can be found in \citet{dibimt}.

\subsection{Implementation Details}

We use pre-trained M2M-100 12B model.\footnote{{\tt last\_checkpoint:}\url{https://github.com/pytorch/fairseq/tree/main/examples/m2m_100}} For quantization, we use Mean Squared Error~(MSE) calibration. For weights, we use default per-channel calibration. In FLORES-101, we use SentencePiece BLEU~(spBLEU) score\footnote{It uses SentencePiece tokenizer with 256K tokens and then BLEU is computed:~\url{https://github.com/facebookresearch/flores}} for the evaluation, as it is shown to be fair for the multilingual comparison~\cite{flores101}. Additionally, we use character \textit{n}-gram F-score~(ChrF)~\cite{popovic-2015-chrf}~\footnote{sacrebleu 1.5.1~\cite{post-2018-call} with ChrF3.} metric to compare compressed models with M2M-100 model. We evaluate our compressed models on language pairs in which M2M-100 12B model~\cite{m2m-100} has reasonable\footnote{Specifically, we choose language pairs, in which M2M-100 12B model has a spBLEU score higher than 12. More details are provided in Appendix~\ref{app:select_pairs}.} performance. This leaves us with 3,763 language directions. All experiments are computed on 2 NVIDIA A100-40GB GPUs.
\section{Results and Discussion}
\label{sec:result}

\subsection{Compression Impact Across Languages}

\begin{table}
\centering
\begin{adjustbox}{width=0.9\linewidth}
\begin{tabular}{lcc}

Resource Type & Criterion & No. Languages \\
\hline
Very-Low & $|L| \leq 100k$ & 16 \\
Low & $100k < |L| \leq 1M $ & 40 \\
Medium & $1M < |L| \leq 100M $ & 38 \\
High & $100M < |L|$ & 7 \\
\hline
\end{tabular}
\end{adjustbox}
\caption{\label{tab:langdist} Distribution of lang. in FLORES-101 based on amount of available data to/from English~($|L|$).}

\end{table}

\begin{figure}
  \centering
  \includegraphics[width=\linewidth]{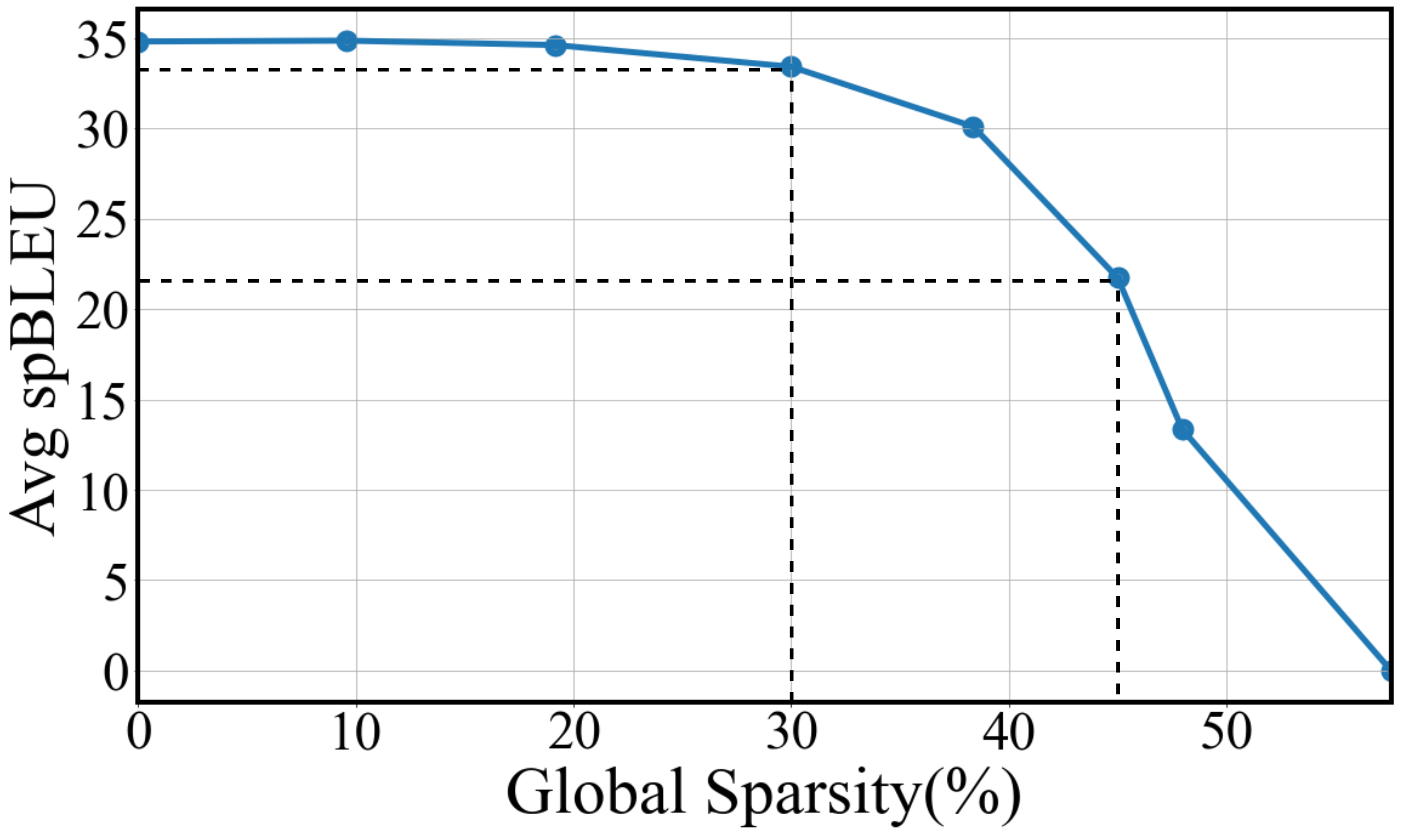}
  \caption{Average spBLEU score for different sparsity ratios on 9 FLORES-101 language pairs, selected from all pairwise combinations of "low", "medium", and "high" language resource categories.}
  \label{fig:sparsity_select}
\end{figure}

\begin{figure*}
\centering
\begin{adjustbox}{width=0.9\linewidth}
\begin{subfigure}[b]{0.3\textwidth}
    \centering
    \includegraphics[width=\textwidth]{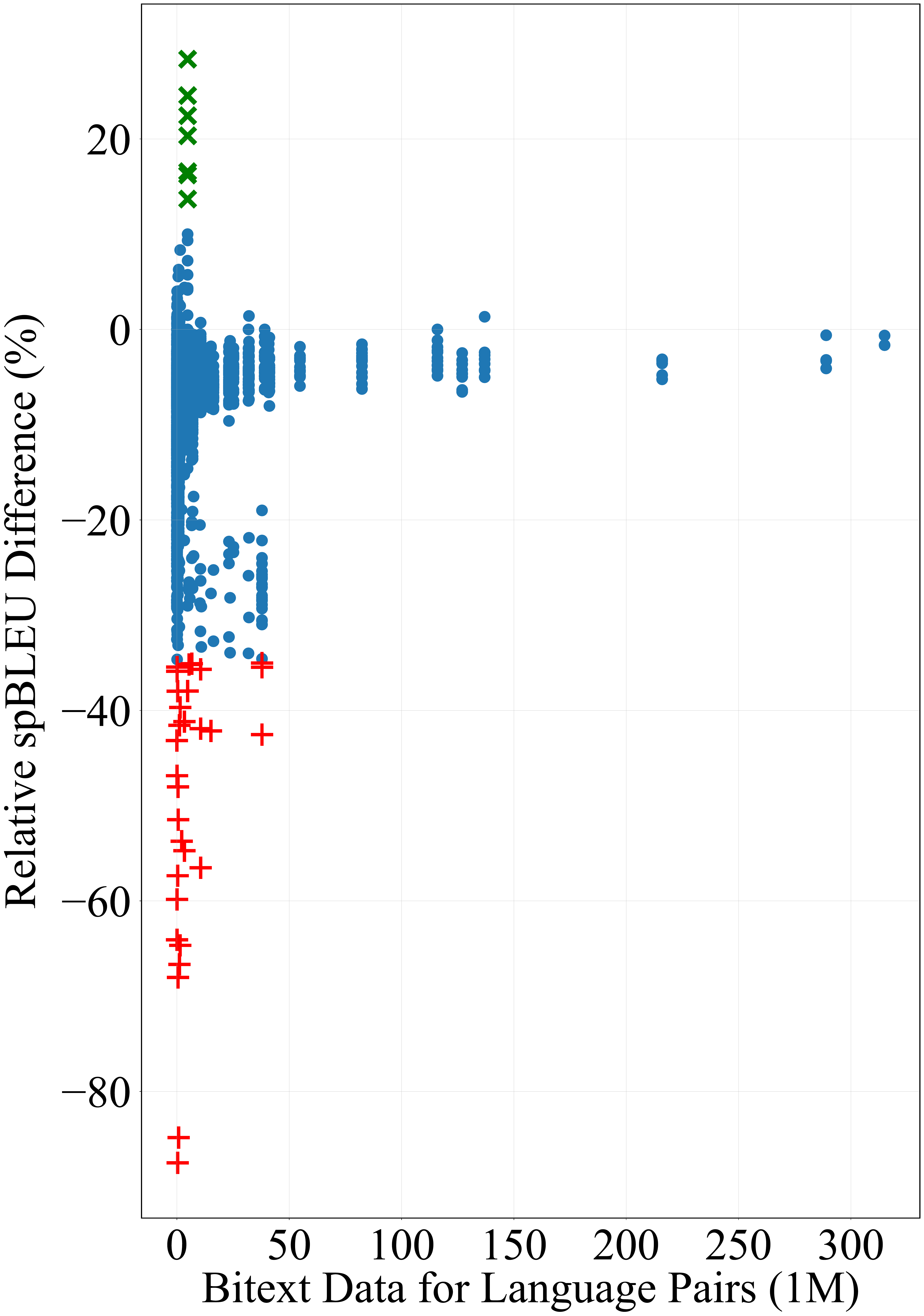}
    \caption{Pruned 30$\%$ Model}
    \label{fig:pruned30}
\end{subfigure}\hfill
\begin{subfigure}[b]{0.3\textwidth}
    \centering
    \includegraphics[width=\textwidth]{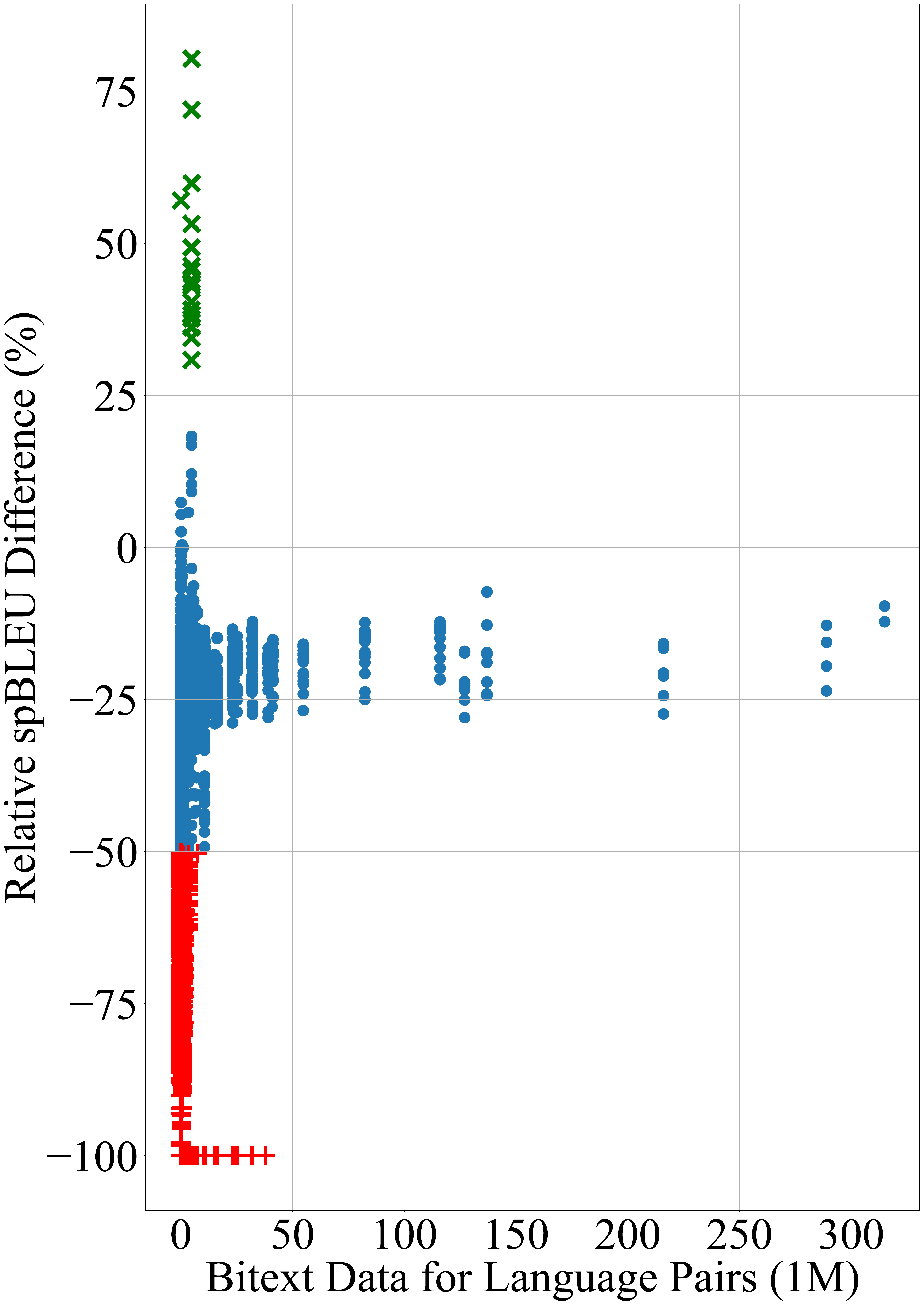}
    \caption{Pruned 45$\%$ Model}
    \label{fig:pruned45}
\end{subfigure}\hfill
\begin{subfigure}[b]{0.3\textwidth}
    \centering
    \includegraphics[width=\textwidth]{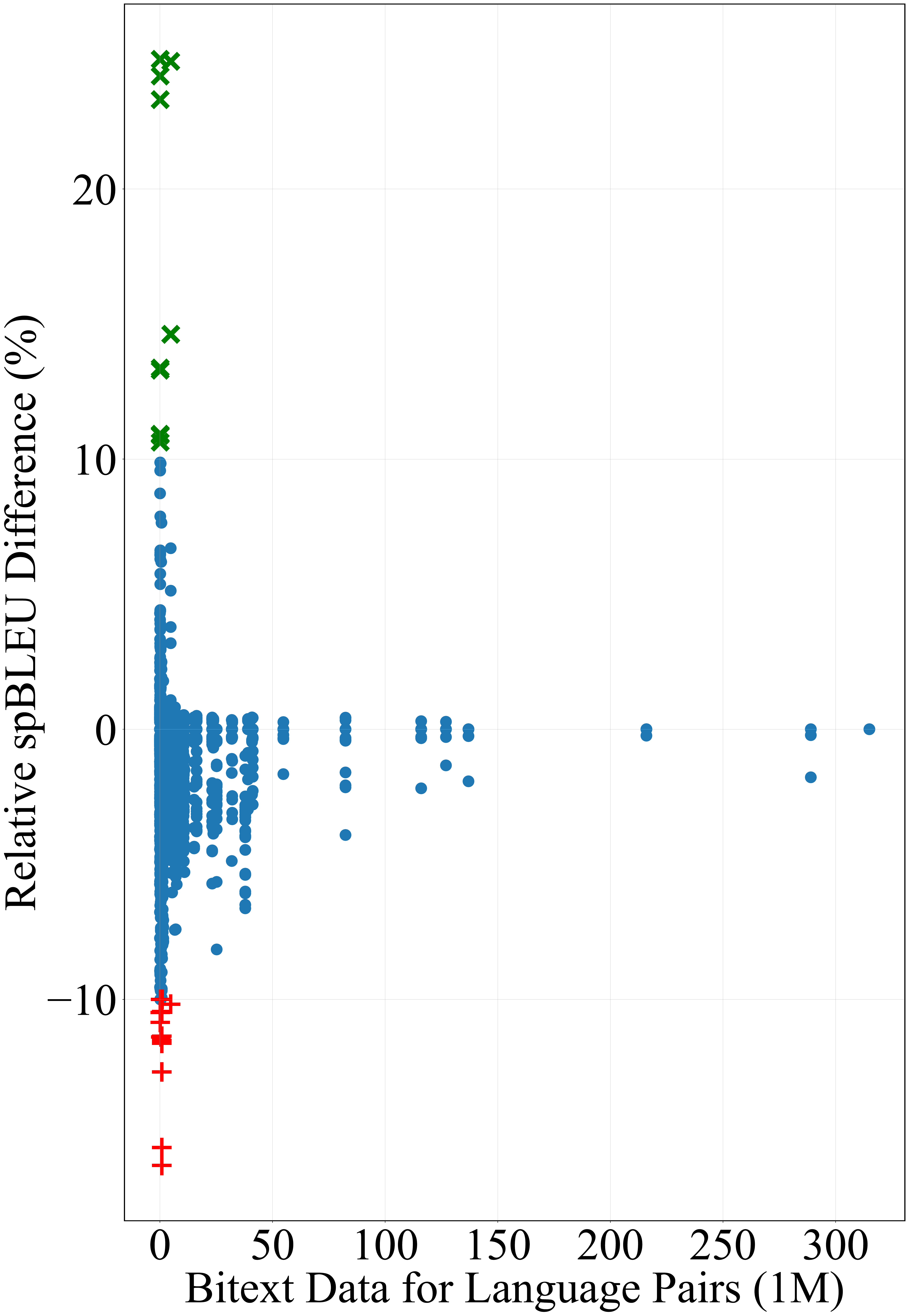}
    \caption{Quantized Model}
    \label{fig:ptq}
\end{subfigure}
\end{adjustbox}
\caption{Relative spBLEU difference~($\%$) between the compressed models and M2M-100 model based on the amount of available Bitext data with English~($\rho_{x,y}$). Green points~("$\times$") are language pairs with significant improvement. Red points~("$+$") correspond to language pairs with a drastic performance drop.}
\label{fig:scatter}
\end{figure*}

\begin{figure}
\centering
\begin{adjustbox}{width=0.9\linewidth}
  \includegraphics[width=\textwidth]{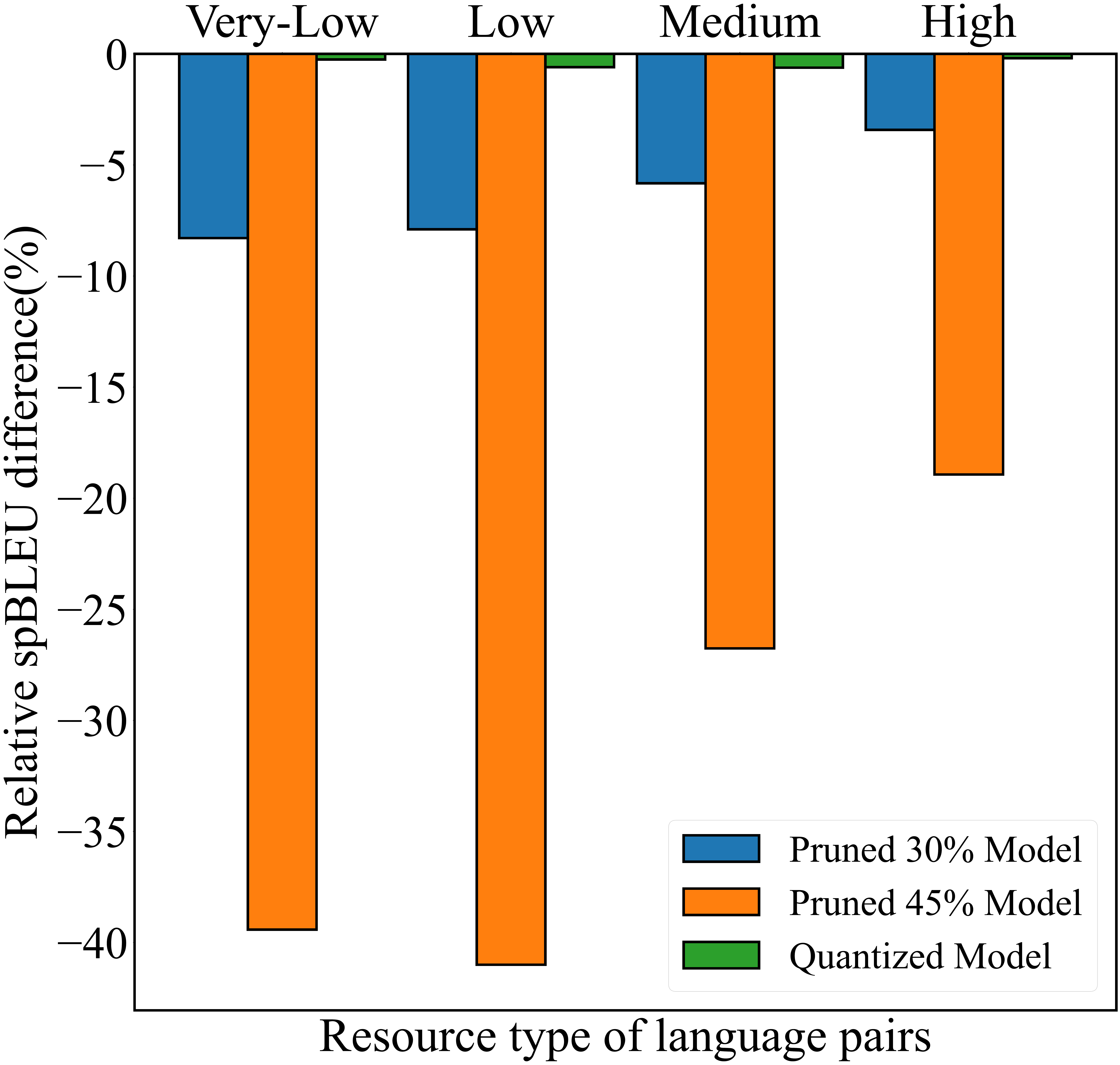}
\end{adjustbox}
\caption{Relative spBLEU difference~($\%$) between the compressed models and M2M-100 model grouped by the resource type of language pairs.\label{fig:type}}
\end{figure}

\paragraph{Language Resource Type.} The true amount of available training data for a language is difficult to estimate, as it relies both on the quality and quantity of the data. Inspired by \citep{flores101}, we classify languages into four categories, based on the amount of available data to/from English. The distribution of language resource types is illustrated in Table~\ref{tab:langdist}.

\paragraph{Magnitude pruning: Sparsity Ratio~($p$) Selection.} Figure~\ref{fig:sparsity_select} shows the average spBLEU score of different sparsity ratios for a subset of language pairs.\footnote{We choose nine language pairs covering all pairwise combinations of "low", "medium", and "high" language categories. A list of this subset is provided in Appendix~\ref{app:sparsity-subset}. } Based on this preliminary analysis, we decide to analyze the model behavior for two sparsity ratios, $30\%$ which is the maximum sparsity ratio for which the compressed model mostly keeps the performance, and $45\%$ for which the performance starts to drop drastically. Therefore, we evaluate the pruned models on sparsity ratios of $30\%$, and $45\%$ for further experiments.

\subsubsection{Main Results}

\begin{table}
\centering
\begin{adjustbox}{width=\linewidth}
\begin{tabular}{lccc}

Model & Memory size & Avg spBLEU & drop($\%$)\\
\hline
M2M-100 & 1$\times$ & \textbf{22.44} & - \\
Pruned 30$\%$ M2M-100 & 0.7$\times$ & \textbf{20.95} & 6.6 \\
Pruned 45$\%$ M2M-100 & 0.55$\times$ & 15.12 & 32.6 \\
Quantized M2M-100 & 0.25$\times$ & \textbf{22.31}  & 0.6 \\
\hline
\end{tabular}
\end{adjustbox}
\caption{\label{tab:avgbleu} Memory size and average spBLEU score of M2M-100, and compressed models on FLORES-101.}

\end{table}

Table~\ref{tab:avgbleu}\footnote{We did not report actual inference time as implementation of compression techniques is highly dependent on the device.} illustrates memory footprint and spBLEU scores on FLORES-101 dataset averaged over 3.7k language pairs retained for analysis. Pruned 30$\%$ model suffers from a slight drop in performance, while quantization mostly preserves the same average spBLEU score. Both quantized and pruned 30$\%$ models reduce the memory footprint by 75$\%$ and 30$\%$, respectively. The performance of $45\%$ pruned model drops significantly. In what follows, we check the behavior of each language pair after compression along different criteria.

\paragraph{Amount of Bitext Data.} Figure~\ref{fig:scatter} shows the relative spBLEU performance of compressed models for each language pair $(x,y)$ compared to the M2M-100. The X-axis corresponds to the amount of bitext data with English defined as 
$\rho_{x,y} = min(\rho_x,\rho_y)$
where $\rho_x$ is the amount of Bitext data with English for language $x$. 
For pruned 30$\%$ model, while the average spBLEU score drops by 6.63$\%$~(shown in Table~\ref{tab:avgbleu}), there is a subset of language pairs that drops drastically~(shown as "$+$"). Interestingly, there is a subset of language pairs that get significantly improved after compression~(shown as "$\times$"). For pruned 45$\%$ model, there is also a subset of languages with more than  50$\%$ drop in performance, while the average spBLEU degradation is 32.62$\%$. For the quantized model which preserves almost the same average spBLEU, we see that there is also a set of languages suffering from a significant drop, and others being significantly improved.  The behavior of compressed models in these specific language pairs is further studied in Section~\ref{sec:losing} and \ref{sec:winning}, respectively. 

\paragraph{Resource Type.} We study the performance of the compressed models based on the resource category of language pairs, which is defined as the category of $\rho_{x,y}$ for a pair $x \rightarrow y$. Figure~\ref{fig:type} demonstrates the relative spBLEU drop for each category of the compressed models. For pruning 30$\%$, the relative spBLEU drop is inversely proportional to the amount of training data for different categories, which confirms that pruning disproportionately impacts the performance of under-represented language pairs, while the average performance is near to the base M2M-100 model~(as shown in Table~\ref{tab:avgbleu}). For quantization, we see a much smaller decrease in all language categories. Furthermore, we show that the resource type of the target language is more crucial than the source language,\footnote{Results are provided in Appendix~\ref{app:spbleu:type}.} meaning that the performance of language pairs with "low" and "very-low" target languages drops drastically after the compression.

\begin{figure}
\centering
\begin{adjustbox}{width=\linewidth}
  \includegraphics[width=\textwidth]{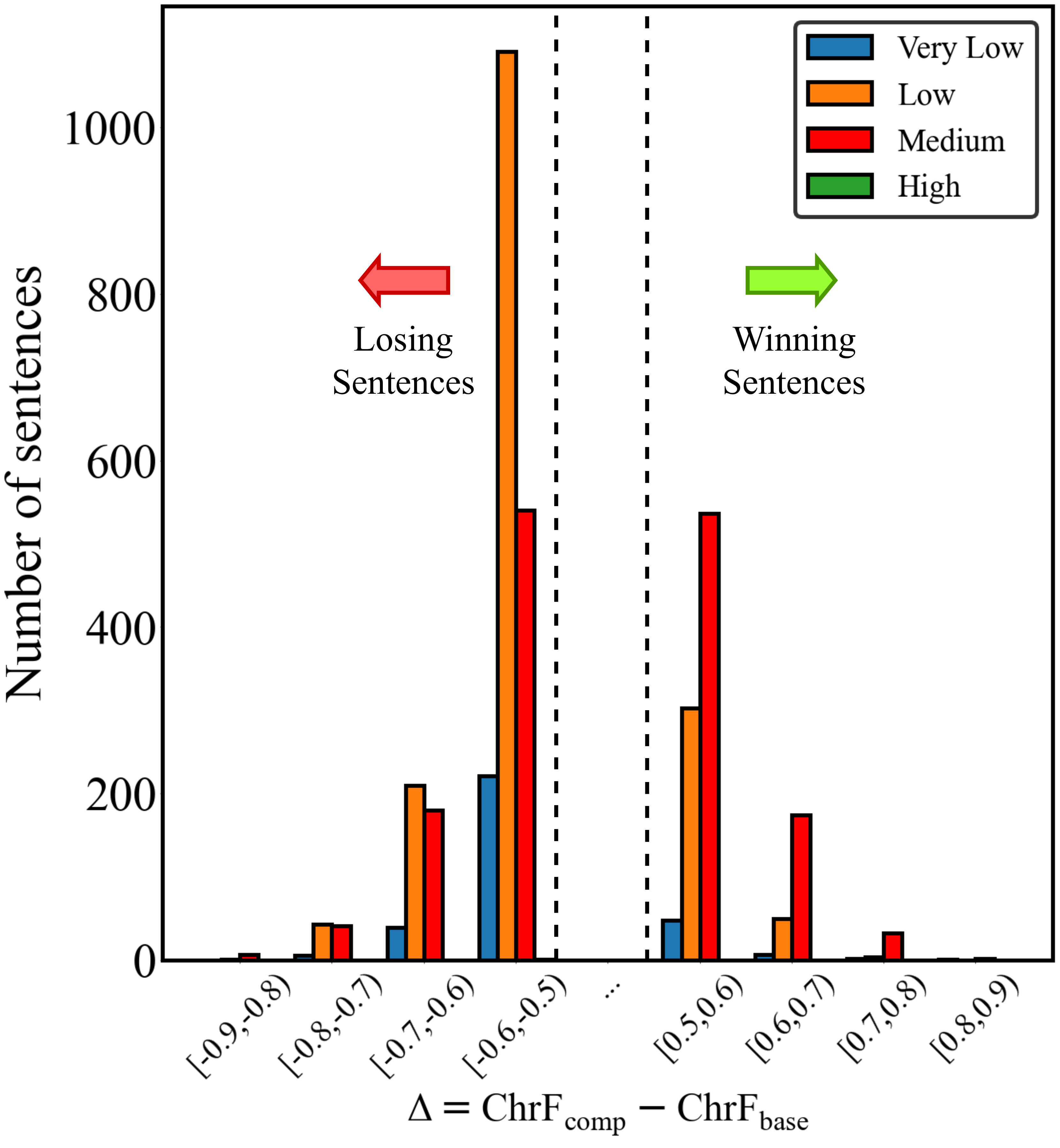}
\end{adjustbox}
\caption{Absolute number of sentences in each language pair category for different $\Delta$ bins.}
\label{fig:chrf}
\end{figure}

\paragraph{ChrF Difference.} For more fine-grained analysis, we perform sentence-level ChrF~\cite{popovic-2015-chrf}\footnote{ChrF demonstrates better correlation with human judgements at sentence-level.} evaluation. We define $\Delta = \mathrm{ChrF}_{\mathrm{comp}} - \mathrm{ChrF}_{\mathrm{base}}$ where $\mathrm{ChrF}_{\mathrm{comp}}$ and $\mathrm{ChrF}_{\mathrm{base}}$ correspond to $\mathrm{ChrF}$ of compressed and baseline models, respectively.  Sentences with $\Delta$ close to zero are less impacted by compression, while those further away from zero are the most impacted (either positively or negatively) by compression.  We define \textit{Losing Pairs} as a set of instances where $\Delta < -0.5$, and  \textit{Winning Pairs} as a set of instances where $\Delta > 0.5$. Thus, identified samples could be seen as an adaptation of \textit{Compression-Identified Exemplars} introduced by \cite{https://doi.org/10.48550/arxiv.1911.05248} for the case of translation.  Figure~\ref{fig:chrf}\footnote{The normalized distribution by the number of instances in each language pair category is provided in Appendix~\ref{app:chrf}.} plots the distribution of sentences from different language pair groups along with the different $\Delta$ bins for these two subsets.~\footnote{Figure~\ref{fig:chrf} belongs to Pruned 30$\%$ model. Complete ChrF calculation~(including $-0.5<\Delta<0.5$) of compressed models for different bins are provided in Appendix~\ref{app:chrf}.}
\begin{table}
\centering
\begin{adjustbox}{width=\linewidth}
\begin{tabular}{l|c|c|c}
Model & Off-T($\%$) \textit{base} & Off-T($\%$) \textit{comp} & Total No. \\
\hline
Pruned 30$\%$ & 5.9 & 13.7(\textbf{+7.8}) & 1,521 \\
Pruned 45$\%$ & 6.4 & 30.3(\textbf{+23.9}) & 10,314 \\
Quantized & 5.2 & 17.5(\textbf{+12.3}) & 268 \\
\hline
\end{tabular}
\end{adjustbox}
\caption{\label{tab:losing} Percentage of off-target translations for M2M-100~(\textit{base}), and compressed models~(\textit{comp}). Last column is the total number of losing sentences~(both on- and off-targets) for each compressed model.}
\end{table}

In the following, we comprehensively analyze the behavior of the model for \textit{Losing Pairs}, and \textit{Winning Pairs}.\footnote{During the preliminary analysis we have identified languages for which M2M-100 training data contains two different scripts (e.g. Cyrillic and Latin), while FLORES-101 dataset provides one script for the evaluation. To fairly analyze the effect of compression, we exclude sentences that refer to these languages. A list of them is provided in Appendix~\ref{app:flores-two-script}.}

\subsubsection{Analysis of Losing Pairs}
\label{sec:losing}
As shown in Figure~\ref{fig:chrf}~(left side), losing pairs belong to very-low, low, and medium-resource languages, that are mostly under-represented subsets during training.\footnote{Normalized distribution in Appendix~\ref{app:chrf} follows same trend.}
We manually inspected some of the translations from the losing pairs sets and we have identified 2 main reasons for the drop in performance which are \textit{off-target translations} (translation in the wrong target language) and \textit{hallucinations}. In what follows we attempt to quantify these two phenomena. 

\paragraph{Off-Target.} We use FastText language identifier~\cite{joulin2016fasttext,joulin2016bag} to predict the languages of reference and the translated sentences. Table~\ref{tab:losing} shows the total number of losing sentences and percentage of off-target translations for both baseline and compressed models.\footnote{We exclude sentences where the predicted reference language ids are not matched with gold reference languages.} As the sparsity increases, the compressed model predicts more off-target translations~(7.8$\%$ and 23.9$\%$ increase from baseline). Quantization also increases the percentage of off-target translation by 12.3$\%$. 
\begin{figure}
\centering
\begin{subfigure}{.45\linewidth}
\centering
            \includegraphics[width=\textwidth,height=1.12\textwidth]{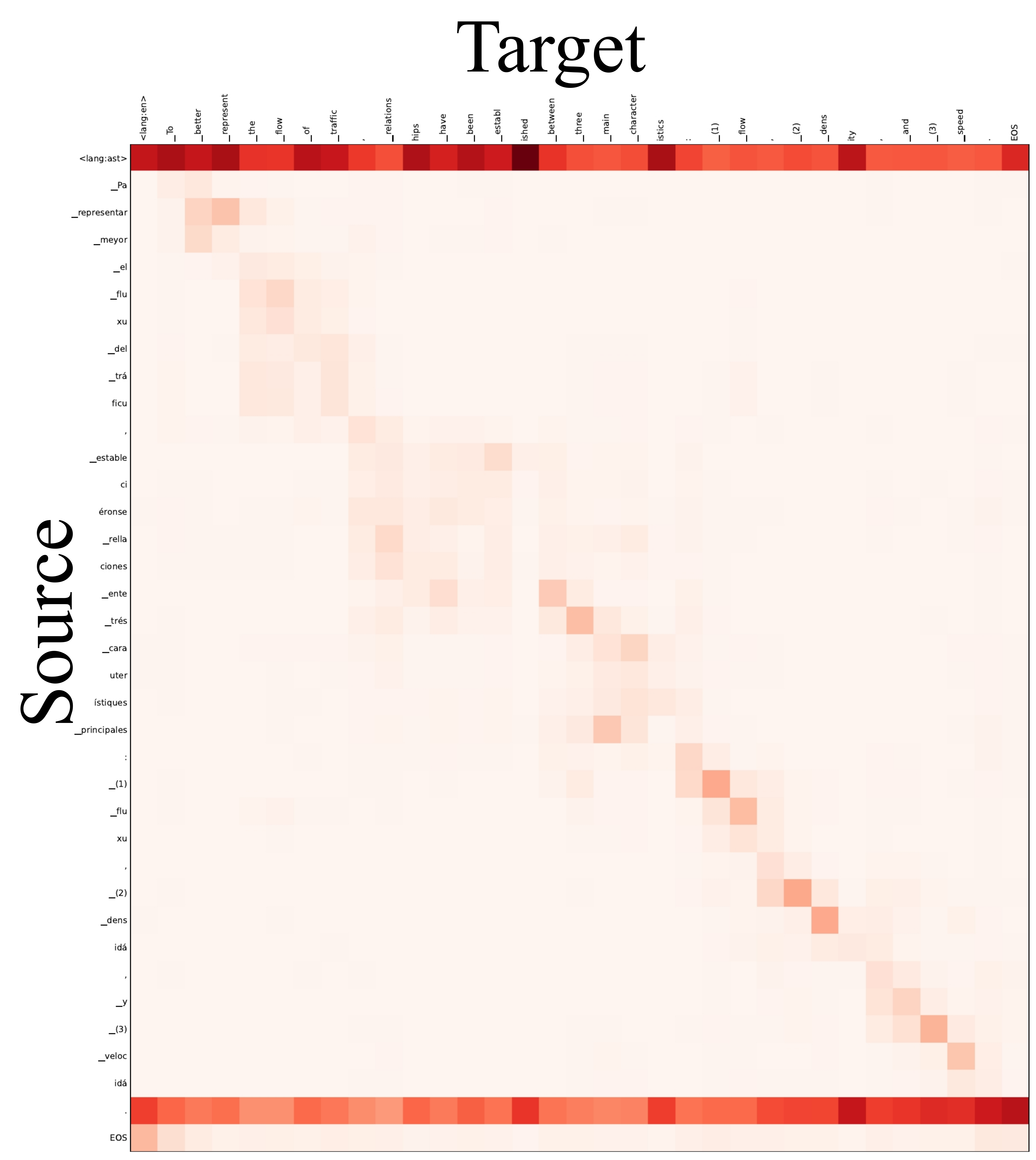}
            \caption{M2M-100 Model}
            \label{fig:lose_base}
\end{subfigure}%
\begin{subfigure}{.45\linewidth}
\centering
            \includegraphics[width=\textwidth,height=1.12\textwidth]{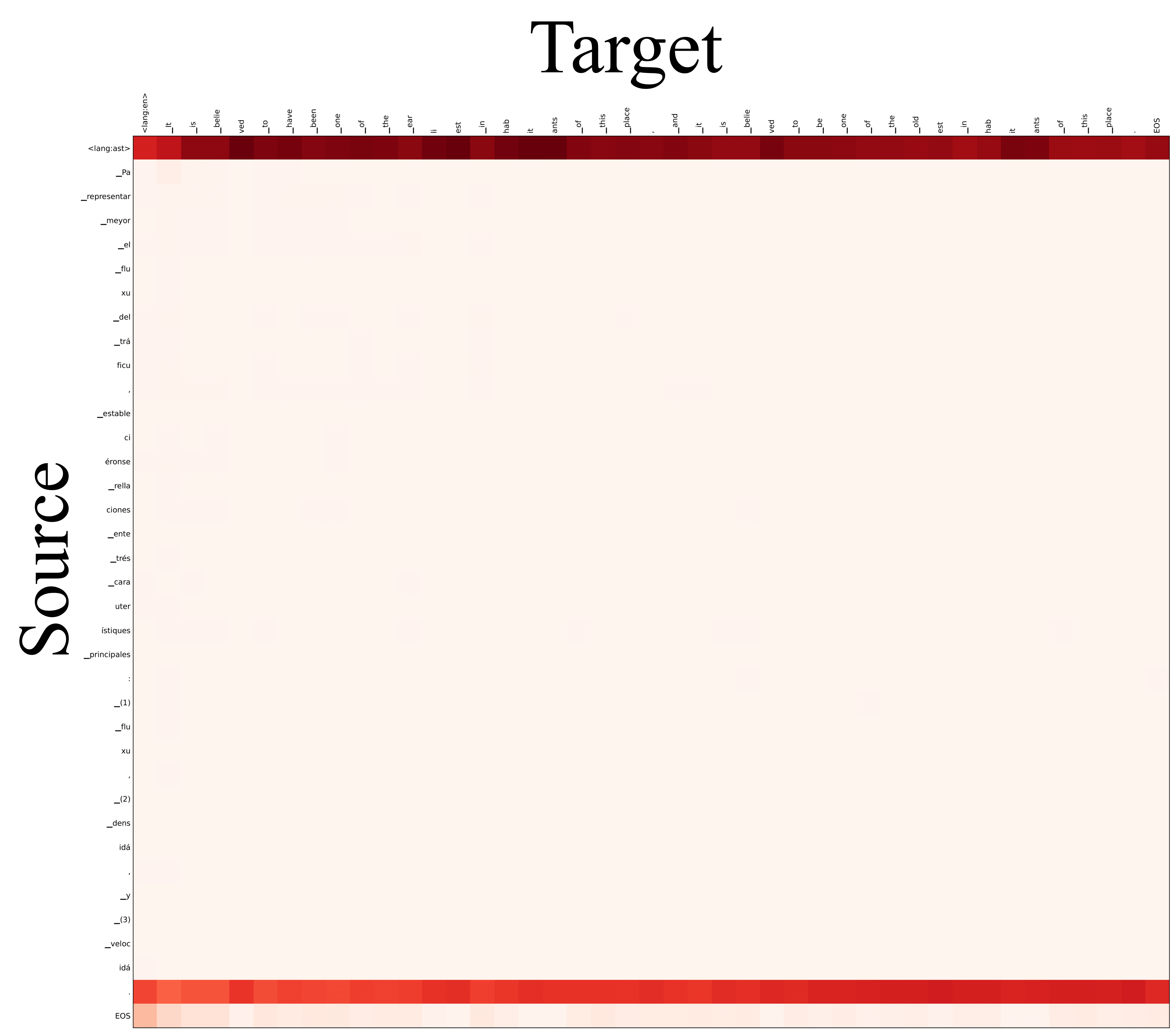}
            \caption{Compressed Model}
            \label{fig:lose_pruned30}
\end{subfigure}\\[1ex]
\begin{subfigure}{0.9\linewidth}
            \begin{minipage}[b]{\linewidth}
            \centering
            \begin{tabularx}{\textwidth}{|c|X|}
            \hline
            \textbf{\small{Reference}} & \small{To better represent traffic flow, relationships have been established between the three main characteristics: (1) flow, (2) density, and (3) velocity.} \\ \hline
            \textbf{\small{M2M-100}} & \small{To better represent the flow of traffic, relationships have been established between three main characteristics: (1) flow, (2) density, and (3) speed.}  \\ \hline
            \textbf{\small{Compressed}} & \small{It is believed to have been one of the earliest inhabitants of this place, and it is believed to be one of the oldest inhabitants of this place.} \\ \hline
            \end{tabularx}
            \caption{Reference and output translations of M2M-100, and compressed models.}
            \label{tab:lose}
            \end{minipage}
\end{subfigure}
\caption{Cross-attention matrices of an on-target losing sentence for the M2M-100 model, and pruned 30$\%$ model. Output translations show the hallucination for the compressed model. Source language is Asturian.}
\label{fig:attn-losing}
\end{figure}
\paragraph{Hallucinations.} It refers to the case, in which a model generates an output unrelated to the source sentence. \citet{lee-etal-2018-character} have shown that the cases of hallucinations have different cross-attention matrices.  
Figure~\ref{fig:attn-losing} shows an example of cross-attention matrices for a losing sentence, where the translation of the compressed model is considered as a hallucination. As expected,  translated tokens ignore the alignment with the source sequence. To quantitatively analyze the hallucination effect on all on-target losing sentences~(excluding off-target translations), we define the relative alignment metric as:
\begin{align}
\begin{split}
\lambda = \frac{\mathrm{var_{comp}}}{\mathrm{var_{base}}}
\end{split}
\end{align}
where $\mathrm{var}$ is defined as:
\begin{align}
\begin{cases}
\mathrm{var} = \frac{1}{|I|.|J|}\sum_{i \in I}\sum_{j \in J}\alpha_{i,j}(\mu_{i}-j)^2 \\
\mu_i = \sum_{j \in J}j.\alpha_{i,j}
\end{cases}
\end{align}
where $I$ and $J$ correspond to sequences of source and target languages, respectively; $\alpha_{i,j}$ is the attention weight, where we use the average attention over all layers and all attention heads. Inspired by \citet{vig-belinkov-2019-analyzing,kim-etal-2021-multilingual}, the variance~($\mathrm{var}$) is high for cases where the target sequence pays attention to a very small subset of source tokens~(hallucination), while it is low when the cross-attention matrix is near to the diagonal matrix~(approximation of perfect alignment matrix).
Table~\ref{tab:losing-lambda} displays the relative alignment~($\lambda$) metric for different compressed models. As the metric is higher than "1" for compressed models, it confirms that target translations of compressed models contain more hallucinated sentences. Lastly, we provide a list of the most affected language pairs in Appendix~\ref{app:most} for further studies.

\begin{table}
\centering
\begin{adjustbox}{width=0.8\linewidth}
\begin{tabular}{l|c|c}
Model & $\lambda$ & No. On-Target sents \\
\hline
Pruned 30$\%$ & 2.95 & 1,312 \\
Pruned 45$\%$ & 3.01 & 7,192 \\
Quantized & 1.96 & 221 \\
\hline
\end{tabular}
\end{adjustbox}
\caption{\label{tab:losing-lambda} Total number of on-target~(excluding off-target translations) sentences and relative alignment~($\lambda$) metric on losing pair subset.}
\end{table}

\begin{table}
\centering
\begin{adjustbox}{width=0.8\linewidth}
\begin{tabular}{l|c|c}
Model & $\lambda$ & Total No. \\
\hline
Pruned 30$\%$ M2M-100 & 0.42 & 863 \\
Pruned 45$\%$ M2M-100 & 0.15 & 1,455 \\
Quantized M2M-100 & 0.52 & 308 \\
\hline
\end{tabular}
\end{adjustbox}
\caption{\label{tab:winning} The relative alignment~($\lambda$) metric for different compressed models on winning pairs subset.}
\end{table}

\begin{figure}
\centering
\begin{subfigure}{.45\linewidth}
\centering
            \includegraphics[width=\textwidth,height=1.12\textwidth]{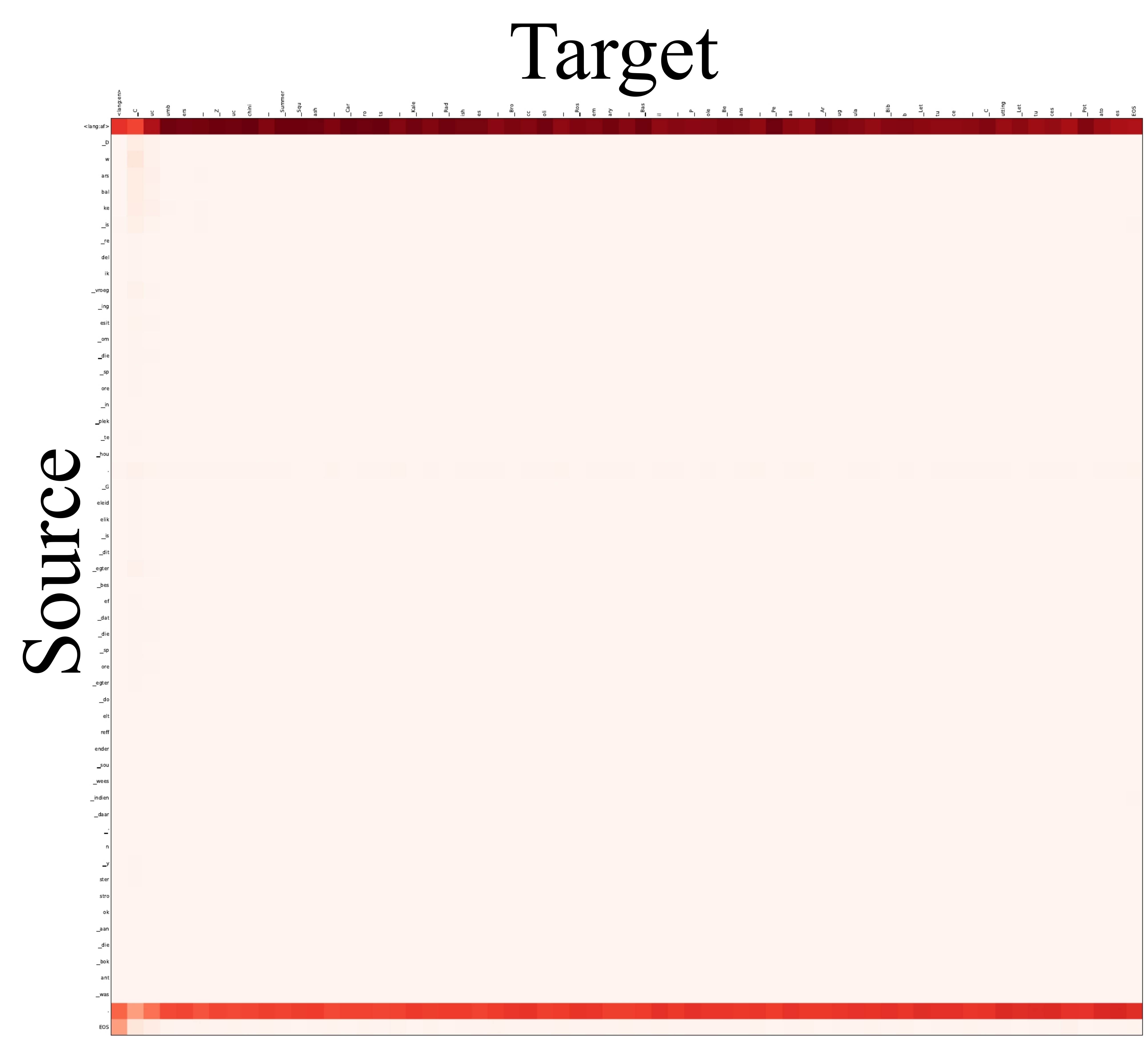}
            \caption{M2M-100 Model}
            \label{fig:win_base}
\end{subfigure}%
\begin{subfigure}{.45\linewidth}
\centering
            \includegraphics[width=\textwidth,height=1.12\textwidth]{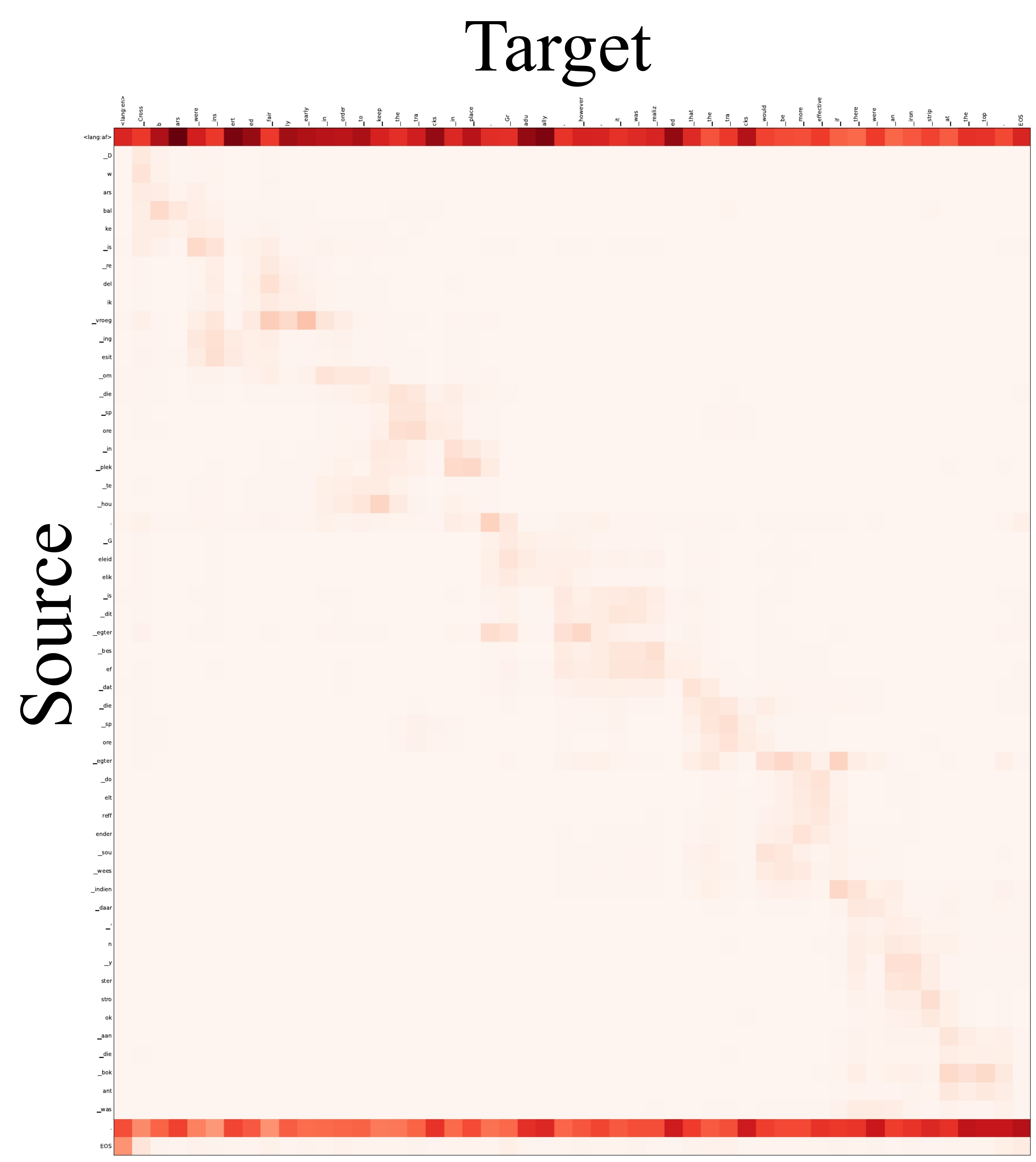}
            \caption{Compressed Model}
            \label{fig:win_pruned30}
\end{subfigure}\\[1ex]
\begin{subfigure}{0.9\linewidth}
            \begin{minipage}[b]{\linewidth}
            \centering
            \begin{tabularx}{\textwidth}{|c|X|}
            \hline
            \textbf{\small{Reference}} & \small{Crossties were introduced fairly early to hold the tracks in place. Gradually, however, it was realised that tracks would be more efficient if they had a stip of iron on the top.} \\ \hline
            \textbf{\small{M2M-100}} & \small{Cucumbers  Zucchini  Summer Squash  Carrots  Kale  Radishes  Broccoli  Rosemary  Basil  Pole Beans  Peas  Arugula  Bibb Lettuce  Cutting Lettuces  Potatoes}  \\ \hline
            \textbf{\small{Compressed}} & \small{Crossbars were inserted fairly early in order to keep the tracks in place. Gradually, however, it was realized that the tracks would be more effective if there were an iron strip at the top.} \\ \hline
            \end{tabularx}
            \caption{Reference and output translations of M2M-100, and compressed models.}
            \label{tab:win}
            \end{minipage}
\end{subfigure}
\caption{Cross-attention matrices of a winning sentence for the M2M-100 model, and pruned 30$\%$ model. Output translations show the hallucination for M2M-100 model. Source language is Afrikaans.}
\label{fig:attn-winning}
\end{figure}

\subsubsection{Analysis of Winning Pairs}
\label{sec:winning}

\begin{figure}
\centering
\begin{adjustbox}{width=\linewidth}
  \includegraphics[width=\textwidth]{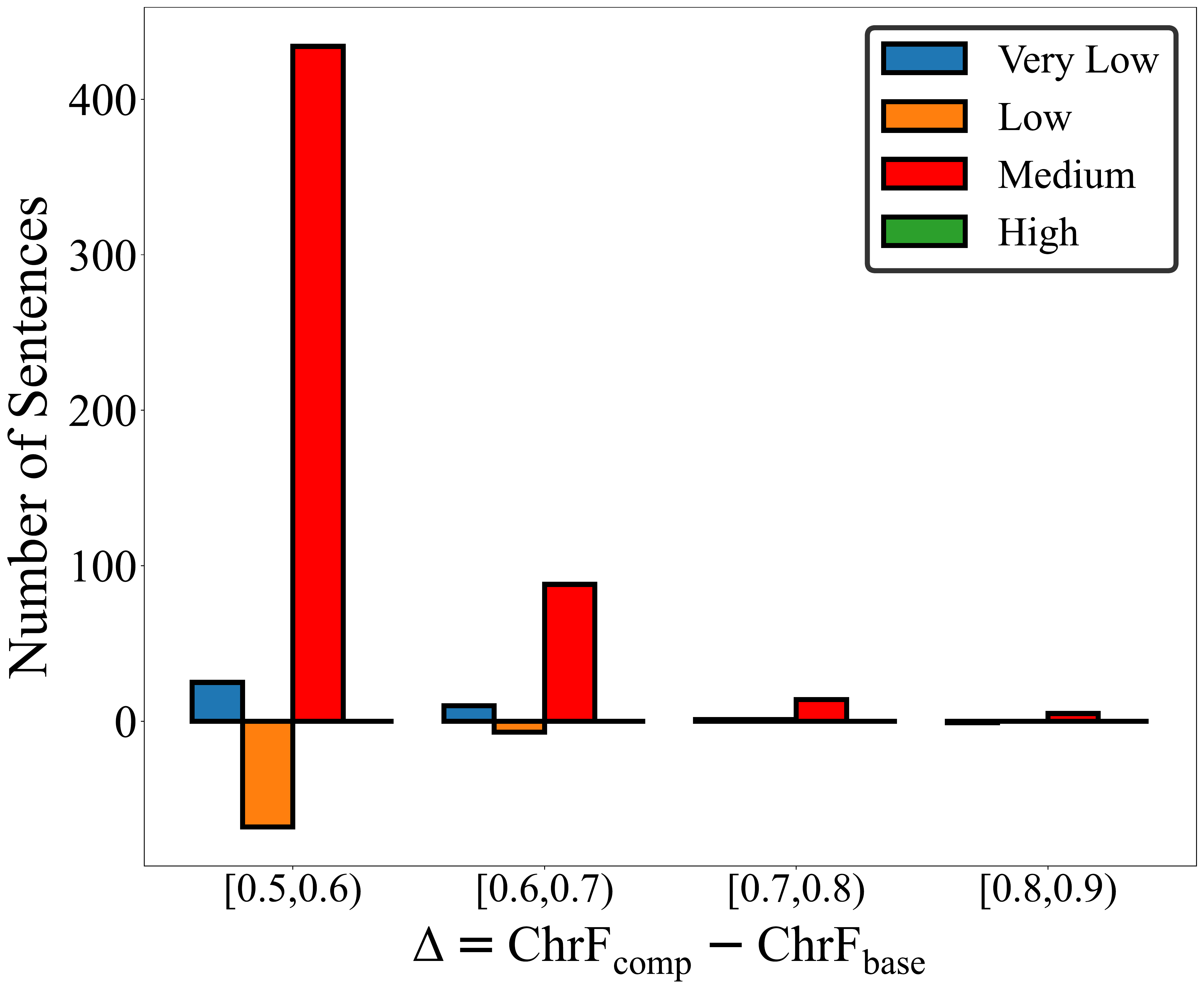}
\end{adjustbox}
\caption{Number of sentences in winning pairs, added to each language category after increasing the sparsity from 30$\%$ to 45$\%$.}
\label{fig:chrf_added}
\end{figure}

When manually inspecting some examples from the translation of winning pairs, we realize that a lot of them are matching cases where the baseline model generates hallucinations,
while the compressed model generates acceptable translations, as shown in Figure~\ref{fig:attn-winning}. We recall that in Figure~\ref{fig:chrf}, most of the winning pairs~(right side) belong to medium-resource languages\footnote{Normalized distribution in Appendix~\ref{app:chrf} shows the same behavior.}, which include a moderate amount of training instances, and could contain some poorly aligned parallel sentences. \citet{raunak-etal-2021-curious} connects the phenomenon of hallucination to the corpus-level noise and suggests that it could also be amplified by back-translation (used for data augmentation to training M2M-100 model). Therefore, the compression seems to remove the memorization of noisy samples, which is more important for medium-resource languages, thus fixing some of the cases of hallucination. 
 In Table~\ref{tab:winning}, we compute the total number of winning sentences, and the relative alignment metric~($\lambda$) for compressed models and M2M-100 model. As $\lambda$ is lower than "1", it confirms that the compression removes the noisy memorization of medium-resource languages, and benefits the generalization of the model. \citet{ahia-etal-2021-low-resource} made a similar observation in the case of bilingual MT models. Interestingly, the number of winning sentences increases as the model gets sparser~(1,455 vs. 863). Figure~\ref{fig:chrf_added} shows that new sentences mostly belong to medium-resource languages. Finally, a list of most winning language pairs is provided in Appendix~\ref{app:most}.

\subsection{Gender Bias Analysis}

We evaluate M2M-100 and our compressed models on MT-Gender benchmark~\cite{stanovsky-etal-2019-evaluating,kocmi-etal-2020-gender}. Inspired by \citet{https://doi.org/10.48550/arxiv.2204.01397}, we use a fairness metric to compare the behavior of compressed models on male and female subsets:
\begin{align}
\begin{split}
\psi = \frac{f_m-f_f}{f_m+f_f}
\end{split}
\label{eq:mtgender1}
\end{align}
where $f_m$, and $f_f$ refer to F1 scores of male and female, respectively. if $\psi$ is near zero, then the model is not biased toward any gender, however, $\psi$ values of +1 or -1 mean that the model is highly biased toward male or female, respectively. We extend the fairness metric to pro- and anti-stereotypical subsets as follows:\footnote{Pro-stereotypical sentences refer to samples that context and occupation match~(e.g. The carpenter stopped the housekeeper and helped her.) while anti-stereotypical subset contains sentences that context and occupation do not match.}:
\begin{align}
\begin{split}
\psi^{\ast} = |\psi_{anti} - \psi_{pro}|
\end{split}
\label{eq:mtgender2}
\end{align}
where $\psi_{pro}$, and $\psi_{anti}$ belong to the fairness metric of pro- and anti-stereotypical sections. Intuitively, if the model has different behaviors in pro- and anti-stereotypical subsets, then it results in increasing the absolute difference of $\psi_{anti}$ and $\psi_{pro}$.\footnote{Proposed metrics are different than simple absolute score difference of \citet{kocmi-etal-2020-gender}, more details in Appendix~\ref{app:mt-gender:metric}.} \\
Average fairness metrics over 10 languages are illustrated in Table~\ref{tab:gender}. Increasing the sparsity ratio results in a more biased model as both $\psi$ and $\psi^{\ast}$ relatively increase +67.2$\%$, and +25.9$\%$. Quantization has less effect on the gender bias as both $\psi$ and $\psi^{\ast}$ negligibly change after applying it. Detailed results for each language are provided in Appendix~\ref{app:mt-gender}. Interestingly, pruning 30\% highly increases the gender bias even for high-resource languages e.g. French and German, while spBLEU is almost the same after the compression.

\begin{table}
\centering
\begin{adjustbox}{width=\linewidth}
\begin{tabular}{l|c|c}
Model & $\psi$~($\%$) & $\psi^{\ast}$~($\%$) \\
\hline
Original M2M-100 & 17.36 & 16.51 \\
Pruned 30$\%$ M2M-100 & 21.65~(\textbf{+24.7}) & 19.52~(\textbf{+18.25}) \\
Pruned 45$\%$ M2M-100 & 29.03~(\textbf{+67.2}) & 20.8~(\textbf{+25.9}) \\
Quantized M2M-100 & 18.24~(+5.1) & 15.53~(-5.8) \\
\hline
\end{tabular}
\end{adjustbox}
\caption{\label{tab:gender} Average fairness metrics over languages of MT-Gender~\cite{stanovsky-etal-2019-evaluating}. Numbers in parentheses are the relative score differences between a specific compressed model and M2M-100 model.}
\end{table}

\subsection{Word Sense Disambiguation Benchmark}
\label{sec:dibimt}

\begin{table}
\centering
\begin{adjustbox}{width=\linewidth}
\begin{tabular}{l|c|c|c|c|c}
Model & SFII & SPDI & MFS & MFS$^{+}$ & AVG \\
\hline
Baseline & 77.6 & 71.6 & 52.8 & 87.6 & 72.4 \\
Pruned 30$\%$ & 76.4 & 72.2 & 52.9 & 87.8 & 72.4 \\
Pruned 45$\%$ & 80.2 & 74.8 & 53.4 & 87.8 & 74.1\\
Quantized & 79.5 & 74 & 53.7 & 88.8 & 74 \\
\hline
\end{tabular}
\end{adjustbox}
\caption{\label{tab:wsd} The average semantic bias metrics over languages of DiBiMT~\cite{dibimt}. Last column is the average score of bias metrics for each model.}
\end{table}

In this section, we analyze the impact of the compression on semantic biases by evaluating our models on a multilingual word sense disambiguation benchmark. We first detail metrics used in \citet{dibimt} to measure semantic biases.
\paragraph{Notation.} Given a specific word~($w_i$), $l_{w_i}$ is defined as (lemmatization, Part-of-Speech tag) pair. $\Pi_{L}(l_{w_i})=\{\sigma_1,...,\sigma_n\}$ is the ordered list of synsets according to WordNet's sense frequency~\cite{10.1093/ijl/3.4.235} in language $L$. For instance, it is built as $\{$the act of firing, photograph, drink, ...$\}$ for noun \textit{shot} in English. $C_{l_{w_i}}(\sigma)$ is the index of synset~($\sigma$) in $\Pi_{L}(l_{w_i})$.
\paragraph{SFII} is calculated as the error rate averaged over $C_{l_{w_i}}(\sigma)$ for different positions and words $w_i$. Intuitively, it measures the sensitivity of the model when predicting a sense concerning its corresponding index in $\Pi_{L}(l_{w_i})$.
\paragraph{SPDI} is computed as the average error rate based on polysemy degrees of synsets.
\paragraph{MFS} measures how often the model chooses more frequent senses than the correct one. Given $C_{l_{w_i}}(\sigma)$ for a synset, it is increased once the model predicts a synset~($\sigma'$) with $C_{l_{w_i}}(\sigma') < C_{l_{w_i}}(\sigma)$.
\paragraph{MFS$^{+}$.} It is similar to the MFS metric, but it increases when $C_{l_{w_i}}(\sigma')$ equals to 1. \\
Since metrics are based on the error rate, the lower values show that the model is less biased. \\
Table~\ref{tab:wsd} demonstrates the semantic bias scores, averaged over all languages in DiBiMT~\cite{dibimt}.\footnote{Detailed results are provided in Appendix~\ref{app:dibimt}.} The last column is the average of semantic bias metrics for each model. According to the average bias score, quantized and pruned 45$\%$ models amplify the bias metric by 1.6, and 1.7 points on average, compared to M2M-100, respectively. It confirms that the compression amplifies the semantic bias while keeping almost the same BLEU performance, especially for the quantization~(average BLEU scores are shown in Table~\ref{tab:avgbleu}).

\section{Related Work}
\label{relatedwork}

The first connection between compression and bias amplification has been made by \cite{https://doi.org/10.48550/arxiv.1911.05248,https://doi.org/10.48550/arxiv.2010.03058} in the case of image classification. The same authors proposed an approach to find a subset of the dataset which contains samples that have disproportionately high errors after the compression. 
There is also recent work that analyzes the effect of compression on pre-trained language models~\cite{xu-etal-2021-beyond,lauscher-etal-2021-sustainable-modular,https://doi.org/10.48550/arxiv.2110.08419,ogueji2022intriguing}. Notably,  \citet{de-vassimon-manela-etal-2021-stereotype} demonstrated a higher gender bias in compressed pre-trained language models.
Concerning NMT, \citet{renduchintala-etal-2021-gender} demonstrated that optimization of inference speed up may result in gender bias amplification. To the best of our knowledge, this work is the first in-depth study of the impact of compression on massively multilingual models.  We hope our findings would encourage further research on this topic. 

\section{Conclusion}

We demonstrate the impacts of applying compression methods to the massively Multilingual Machine Translation models by evaluating compressed models on FLORES-101~\cite{flores101}, gender bias benchmark~\cite{stanovsky-etal-2019-evaluating}, and word sense disambiguation benchmark~\cite{dibimt}. We show that while average BLEU drops negligibly, the performance of under-represented language pairs drops drastically. Interestingly,  sparsity improves the performance of some medium-resource language pairs by removing the noisy memorization. By evaluating our compressed models on gender bias and word sense disambiguation benchmarks, we show that the compression amplifies the intrinsic gender and semantic biases, even in high-resource language pairs. We hope our findings could be a starting point to consider the fairness aspects when compressing multilingual models.
\section*{Limitations}

Our compression techniques are limited to post-training quantization, and magnitude pruning without additional fine-tuning due to the huge cost of fine-tuning these massively multilingual models, but future research could extend our analysis to compression methods with additional fine-tuning, e.g. knowledge distillation~\cite{kim-rush-2016-sequence}, training-aware pruning and quantization~\cite{behnke-heafield-2020-losing,zhang-etal-2021-enlivening,yao2022zeroquant}. We analyze our compressed models based on the amount of available training data for each language pair, gender bias, and word sense disambiguation bias. Future research could apply our analysis to other linguistic biases in the machine translation task. 
\section*{Acknowledgement}

Alireza Mohammadshahi is supported by the Swiss National Science Foundation~(grant number CRSII5-180320).

\bibliographystyle{acl_natbib}
\bibliography{emnlp2022}

\newpage
\renewcommand\thesection{\Alph{section}}
\renewcommand\thesubsection{\thesection.\Alph{subsection}}

\begin{appendices}
\onecolumn

\section{Magnitude Pruning Strategy}
\label{app:magn-strategy}

\begin{figure}[!ht]
\centering
  \includegraphics[width=0.7\textwidth]{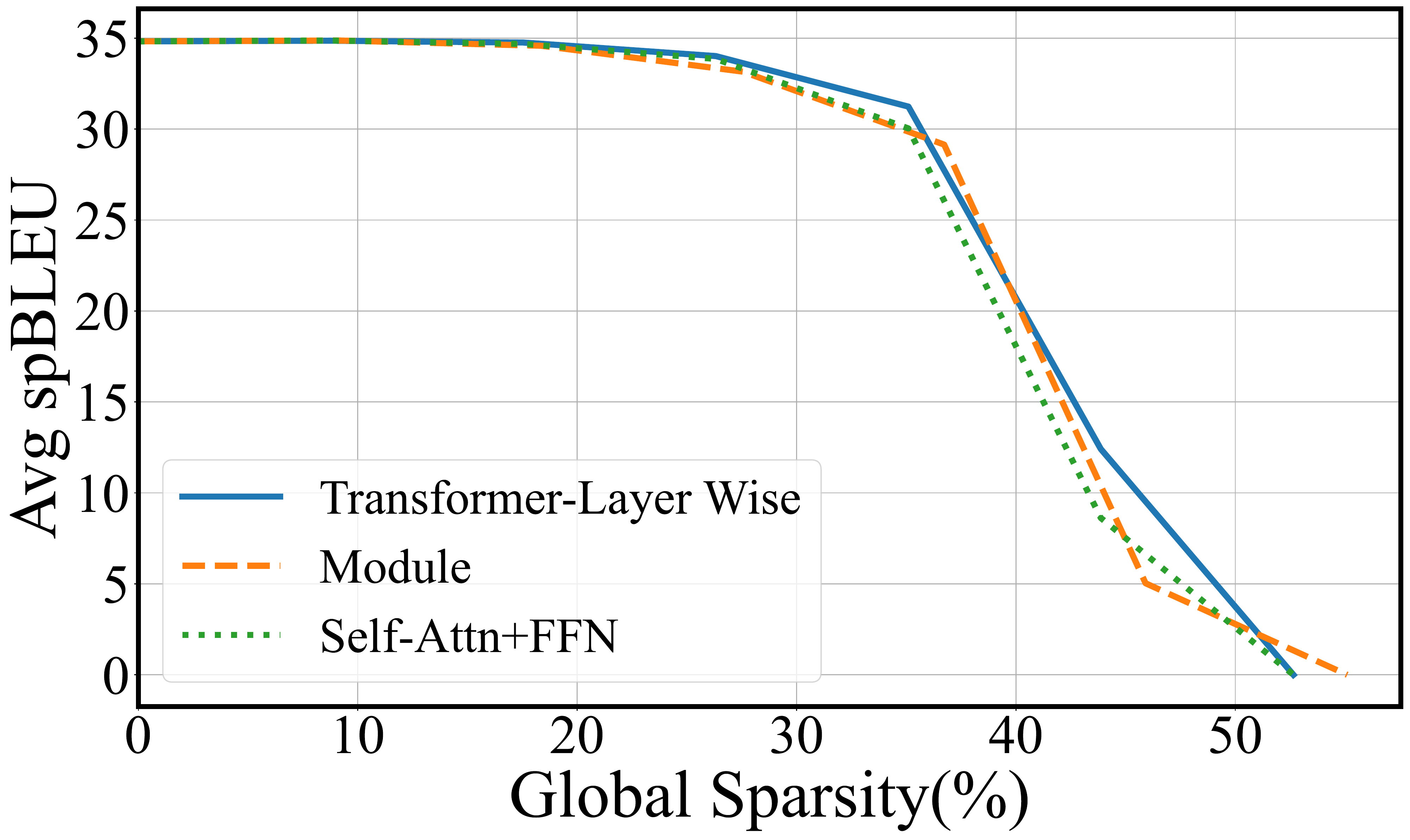}
  \caption{Average spBLEU score of different magnitude pruning strategies on 9 FLORES-101 language pairs, defined in Appendix~\ref{app:sparsity-subset}.\label{fig:app:mgn-str}}
\end{figure}

Figure~\ref{fig:app:mgn-str} shows the performance of pruned models with different pruning strategies. Results illustrate that pruning based on Transformer-layer is slightly better than pruning based on each module of the model, and separate pruning for self-attention and feed-forward Transformer layers.
    
\section{Selection of Language Pairs in FLORES-101}
\label{app:select_pairs}

\begin{figure}[!ht]
  \centering
  \includegraphics[width=0.7\linewidth]{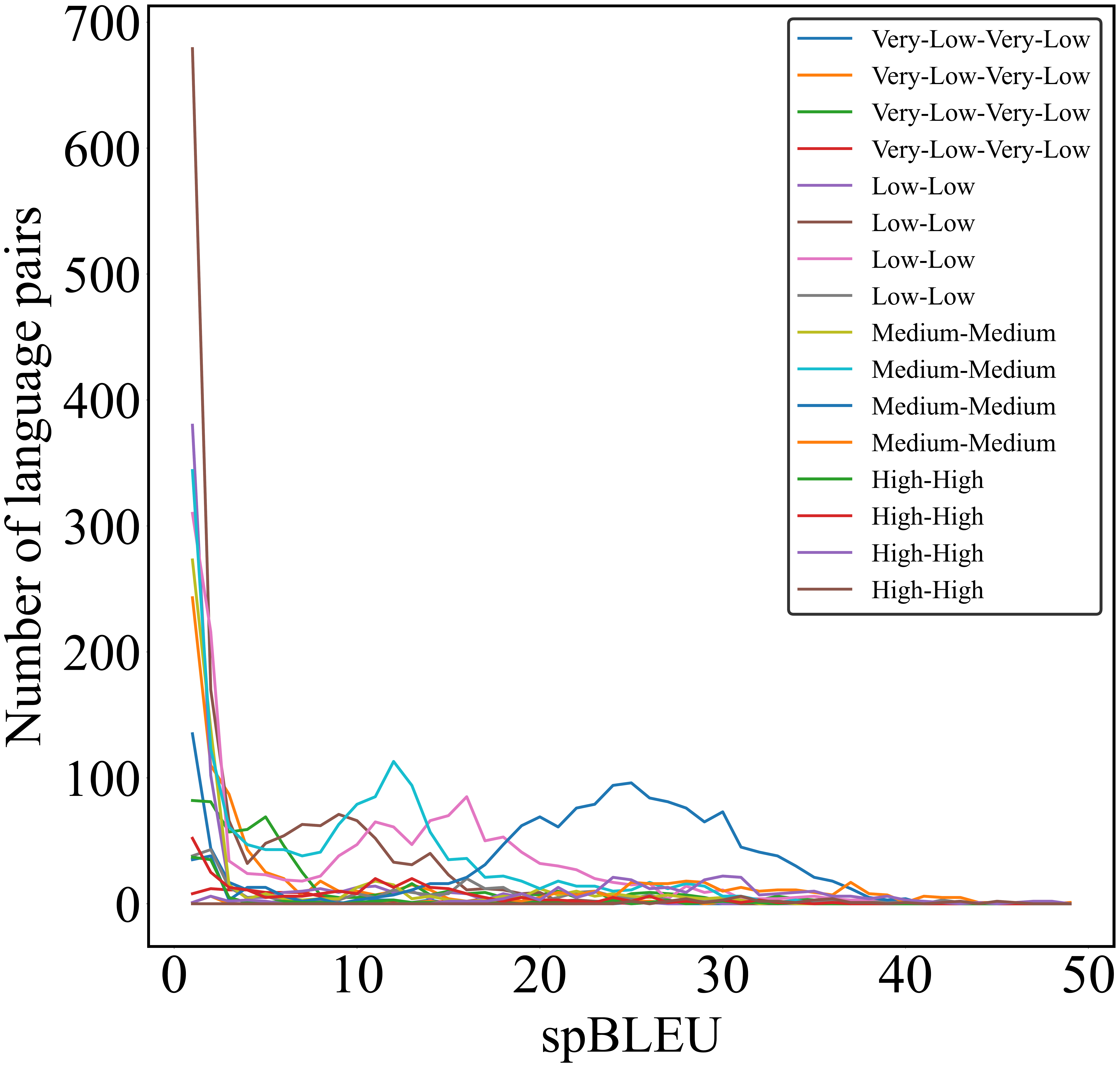}
  \caption{Histogram of number of language pairs based on spBLEU score for different language pair categories.}
  \label{fig:app:select_pair}
\end{figure}

Figure~\ref{fig:app:select_pair} shows the distribution of different language pair categories~(defined in Table~\ref{tab:langdist}) based on spBLEU score of M2M-100 12B model~\cite{m2m-100}. We use 12 spBLEU as the threshold, which is approximately the average score over the  median of different language pair categories.

Table~\ref{tab:app:select_pair} illustrates the number of language pairs in each category after the filtering. 

\begin{table}[!ht]
\centering
\begin{adjustbox}{width=0.5\linewidth}
\begin{tabular}{c|c|c|c|c}

\diagbox{Source}{Target} & Very-Low & Low & Medium & High \\ \hline
Very-Low & 10 & 51 & 157 & 33 \\ \hline
Low & 58 & 164 & 643 & 143 \\ \hline
Medium & 108 & 440 & 1,277 & 257 \\ \hline
High & 23 & 103 & 252 & 39 \\ \hline
\end{tabular}
\end{adjustbox}
\caption{\label{tab:app:select_pair}Number of language pairs in each category after the filtering.}
\end{table}

\newpage
\section{Language Pairs for Selection of Sparsity Ratio}
\label{app:sparsity-subset}

\begin{table}[!ht]
\centering
\begin{adjustbox}{width=0.6\linewidth}
\begin{tabular}{l|c|c}

Language Pair & Resource-Type & M2M-100 spBLEU\\
\hline
 Bosnian-Afrikaans & low-to-low & 29.9\\
 Afrikaans-Bulgarian & low-to-medium & 37.3 \\
 Afrikaans-French & low-to-high & 41.5\\
 Catalan-Asturian & medium-to-low & 29.7 \\
 Danish-Bulgarian & medium-to-medium & 37.8 \\
 Swedish-Spanish & medium-to-high & 27.5 \\
 French-Afrikaans & high-to-low & 30.9\\
 Spanish-Swedish & high-to-medium & 27.5 \\
 English-French & high-to-high & 51.3\\
\hline
\end{tabular}
\end{adjustbox}
\caption{Subset of language pairs used to compute average spBLEU score of Figure~\ref{fig:sparsity_select}. M2M-100 model achieves reasonable performance for all selected pairs as shown in the last column. }
\end{table}

\section{Relative spBLEU based on Resource Type of Target and Source}
\label{app:spbleu:type}
~~
\begin{figure}[!htb]
\centering
\begin{subfigure}[b]{0.48\textwidth}
    \centering
    \includegraphics[width=\textwidth]{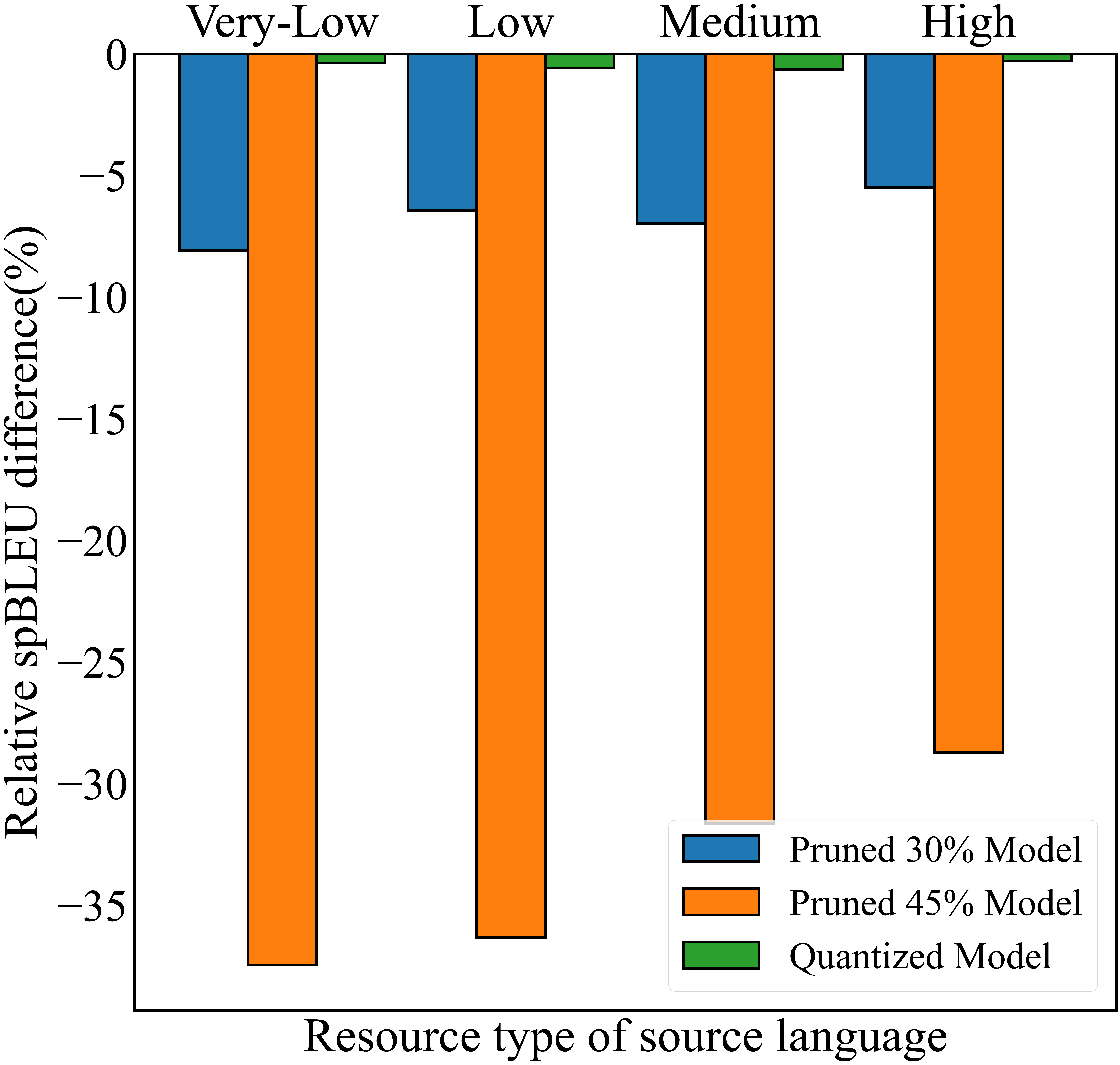}
    \caption{Source Resource Type}
\end{subfigure}\hfill
\begin{subfigure}[b]{0.48\textwidth}
    \centering
    \includegraphics[width=\textwidth]{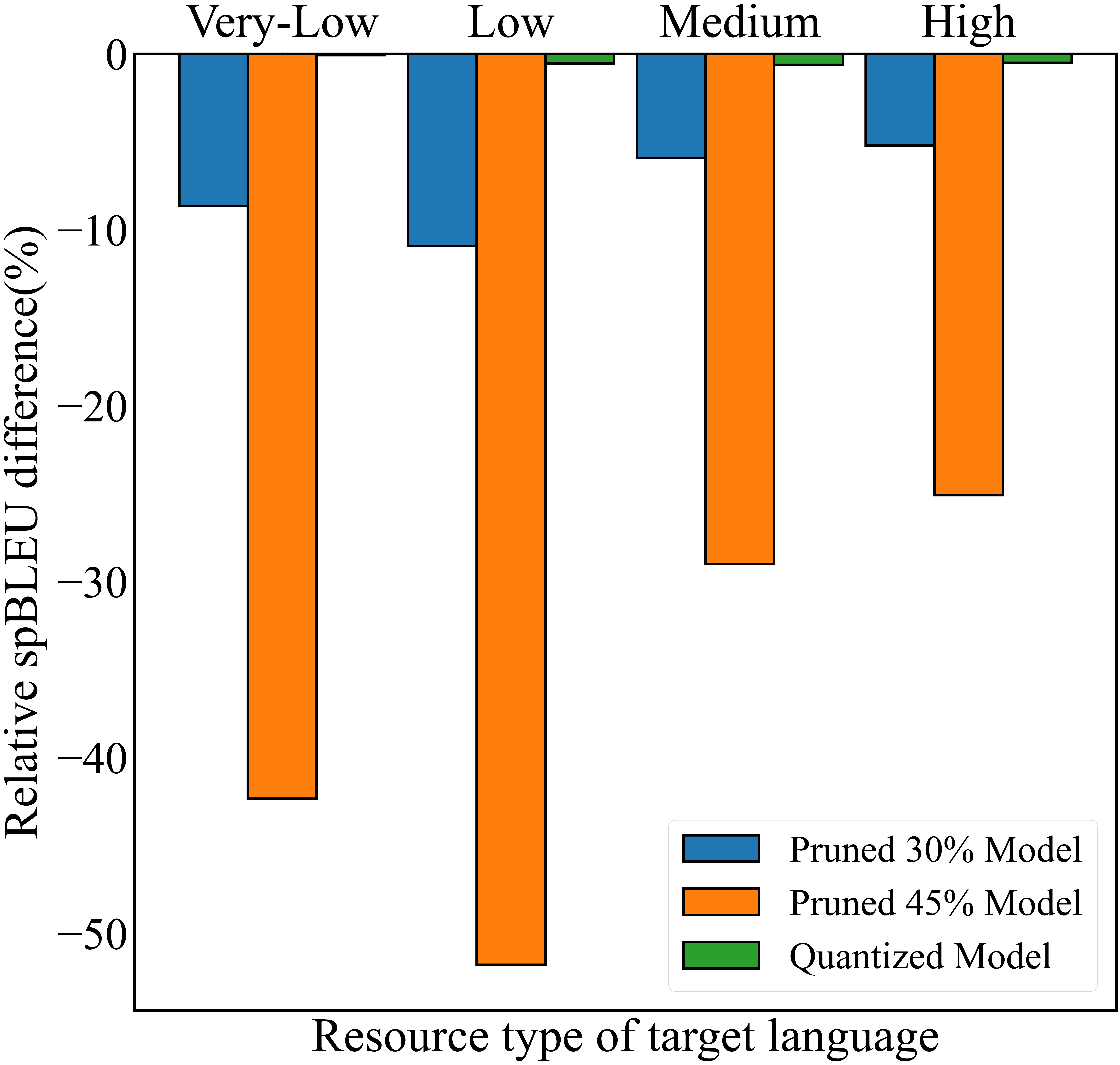}
    \caption{Target Resource Type}
\end{subfigure}
\caption{Relative spBLEU difference~($\%$) between compressed models and M2M-100 model grouped by the resource type of source or target languages.}
\label{fig:app:bleu_type}
\end{figure}
~~~
\newpage
~~~
\newpage
\section{ChrF Difference Analysis}
\label{app:chrf}

\subsection{Pruned 30$\%$ Model}
~~

\begin{figure}[!htb]
\begin{subfigure}{.5\linewidth}
\centering
\includegraphics[width=\textwidth]{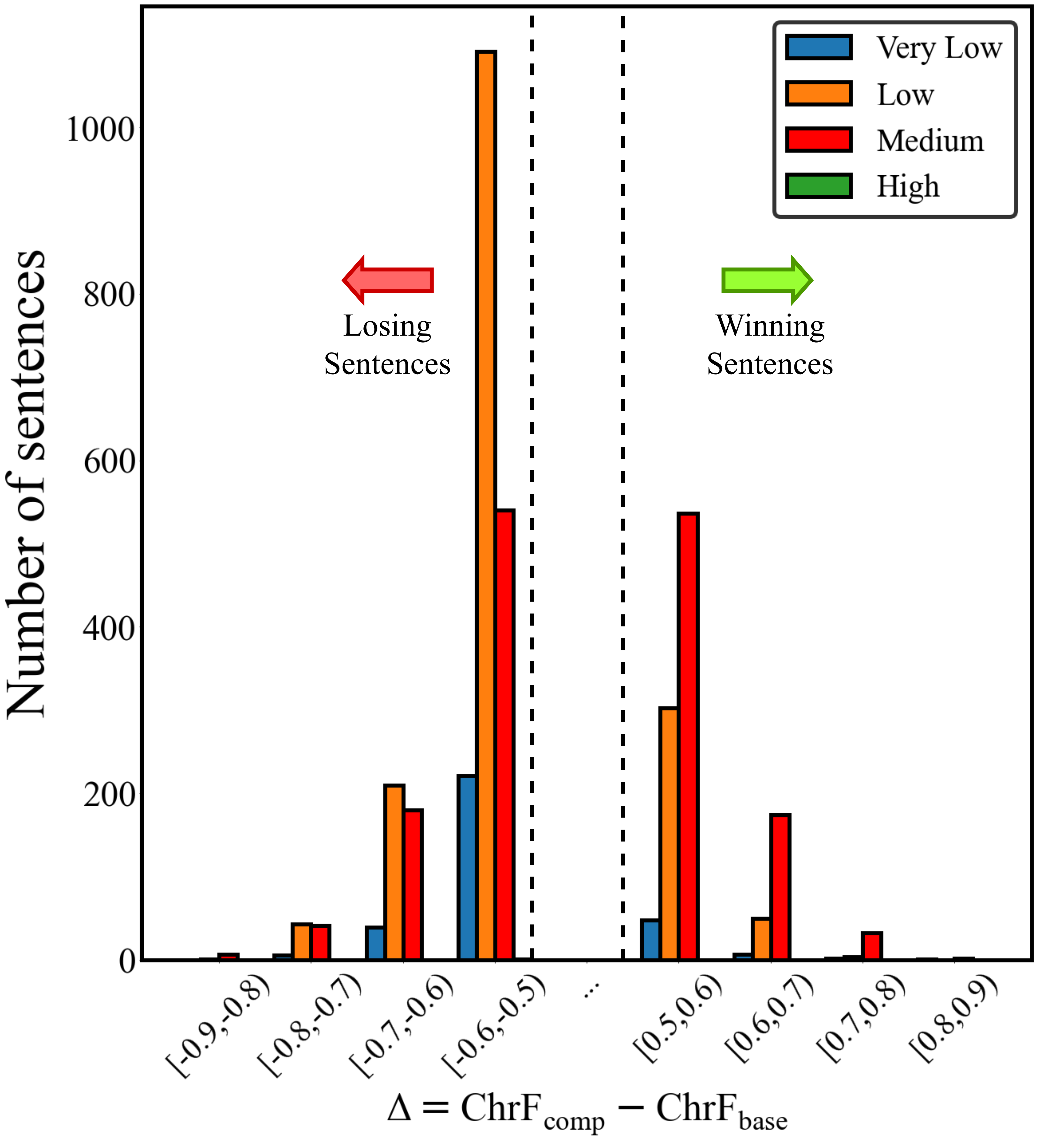}
\caption{Absolute number of sentences.}
\end{subfigure}%
\begin{subfigure}{.5\linewidth}
\centering
\includegraphics[width=\textwidth]{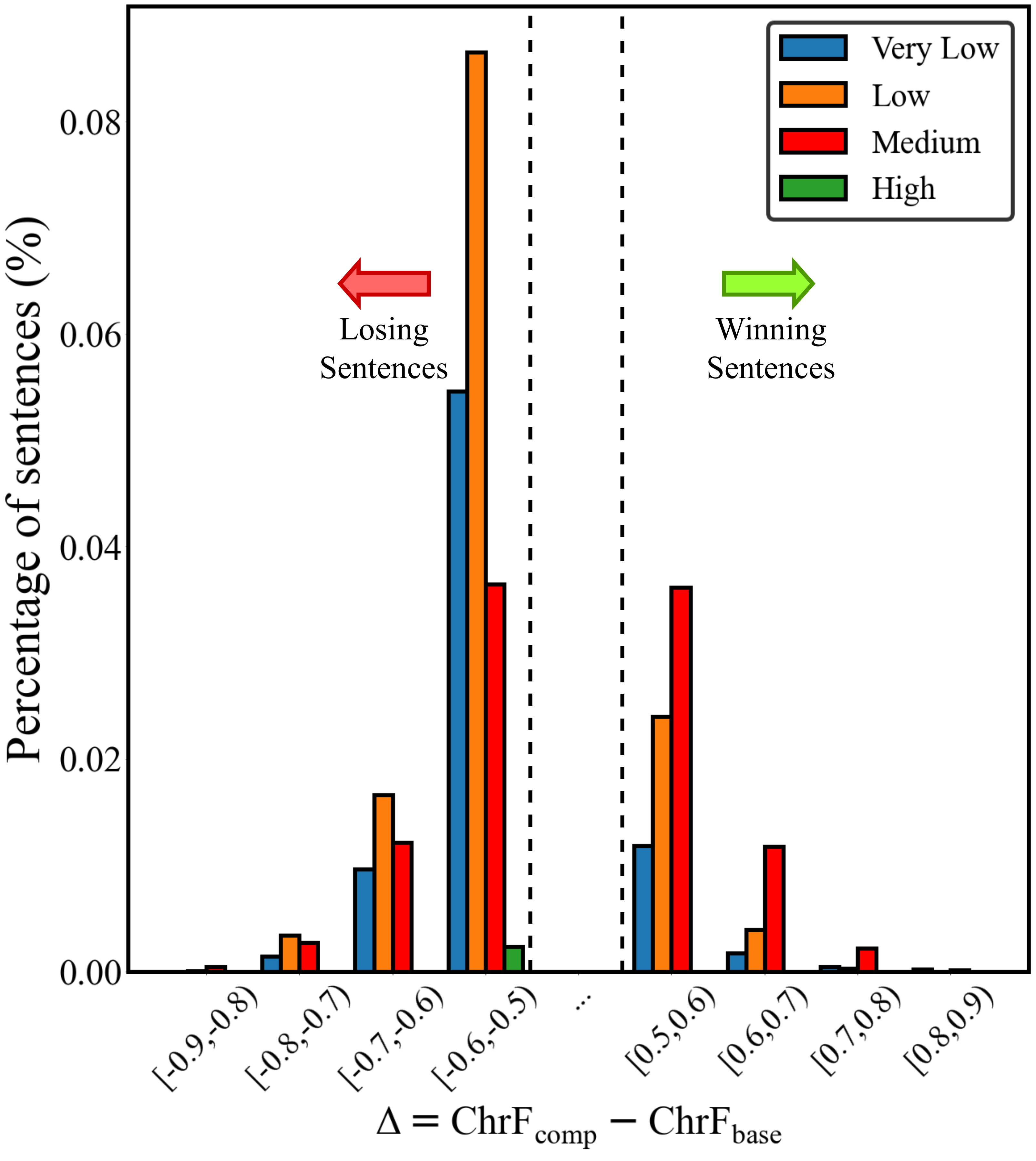}
\caption{Normalized distribution of sentences.}
\end{subfigure}\\[1ex]
\begin{subfigure}{\linewidth}
\centering
\includegraphics[width=\textwidth]{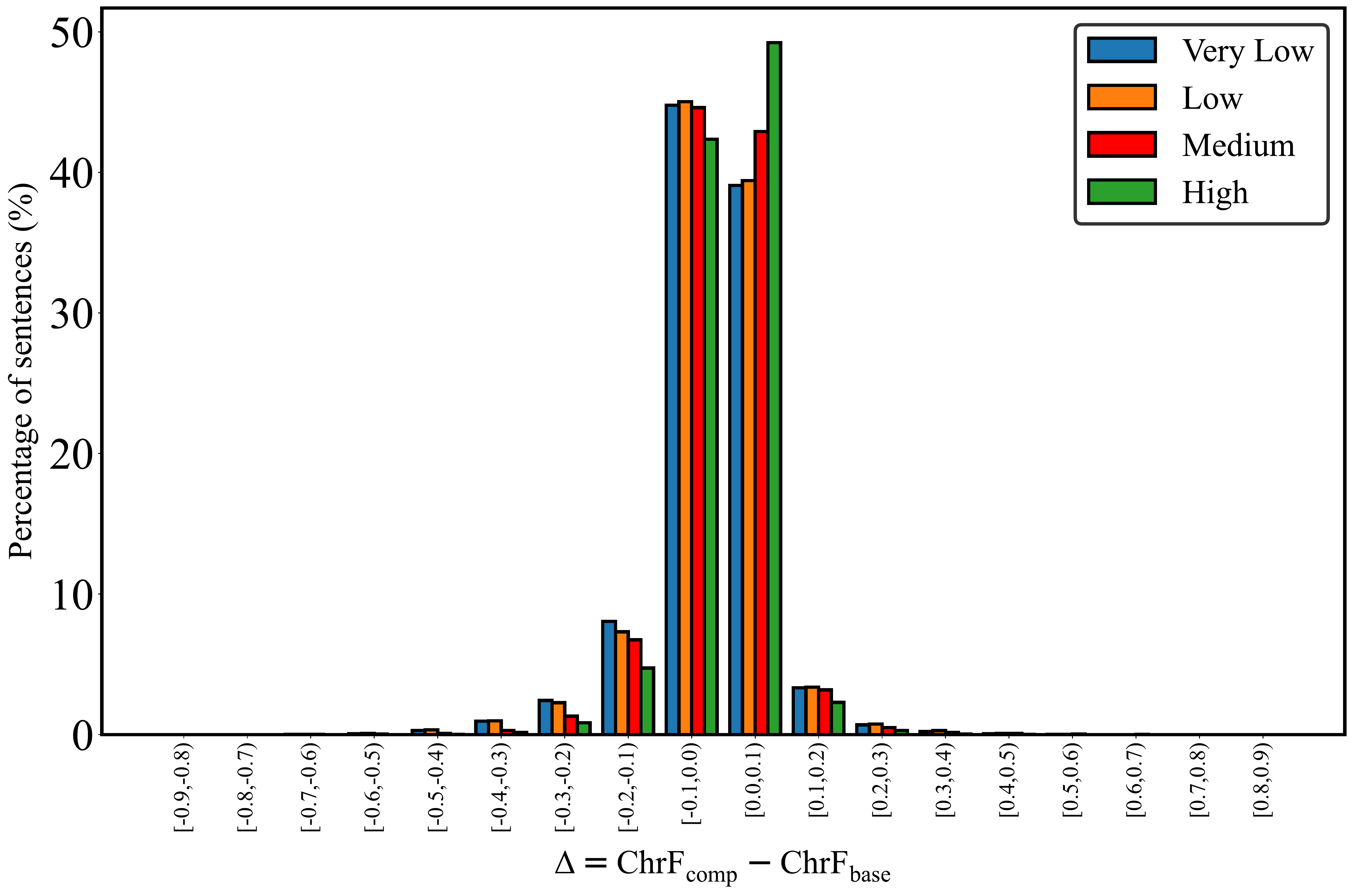}
\caption{Normalized distribution of sentences in each bin for different categories.}
\end{subfigure}
\caption{ChrF analysis of pruned 30$\%$ M2M-100 model.}
\end{figure}

\subsection{Pruned 45$\%$ Model}
~~

\begin{figure}[!htb]
\begin{subfigure}{.5\linewidth}
\centering
\includegraphics[width=\textwidth]{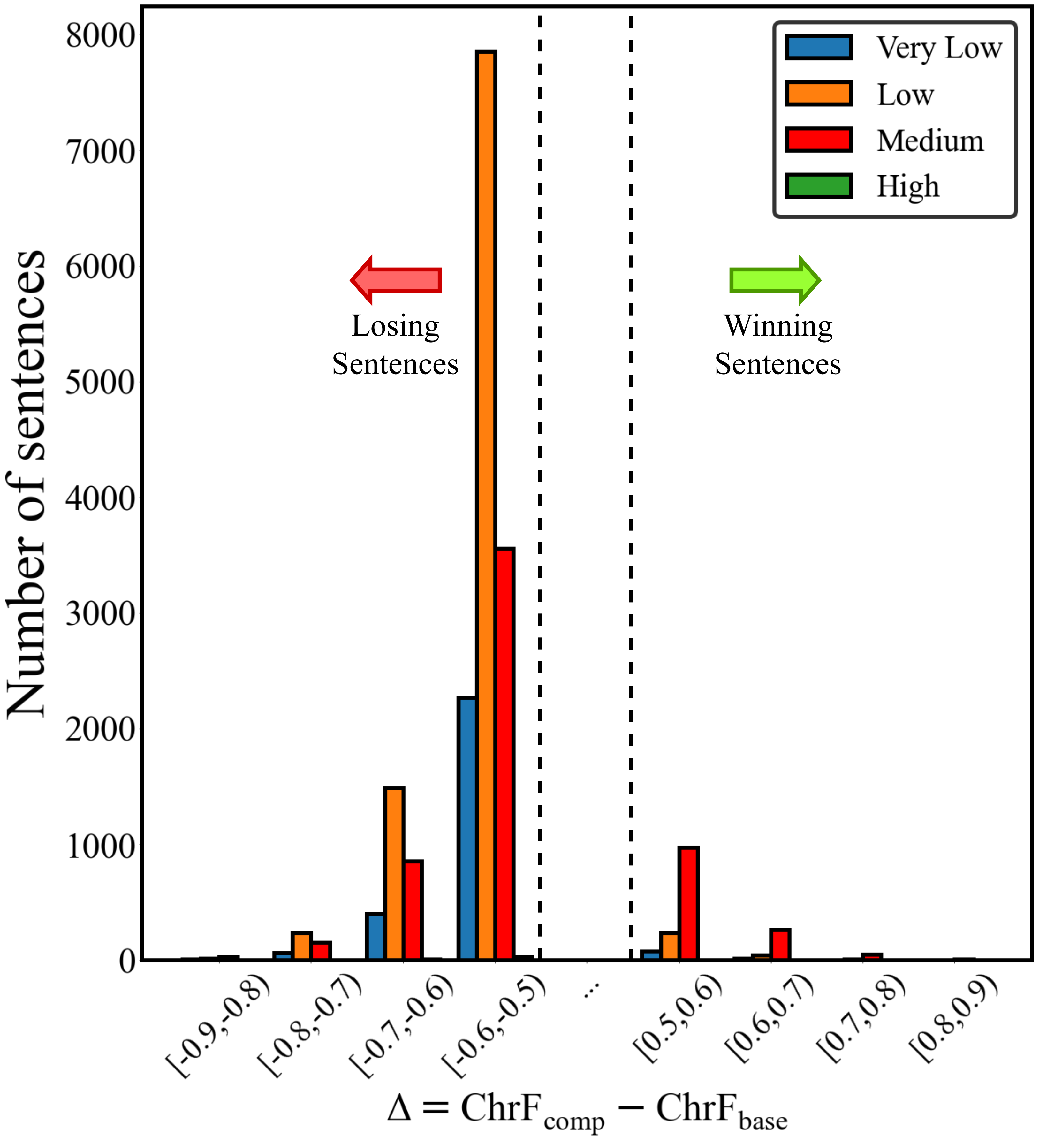}
\caption{Absolute number of sentences.}
\end{subfigure}%
\begin{subfigure}{.5\linewidth}
\centering
\includegraphics[width=\textwidth]{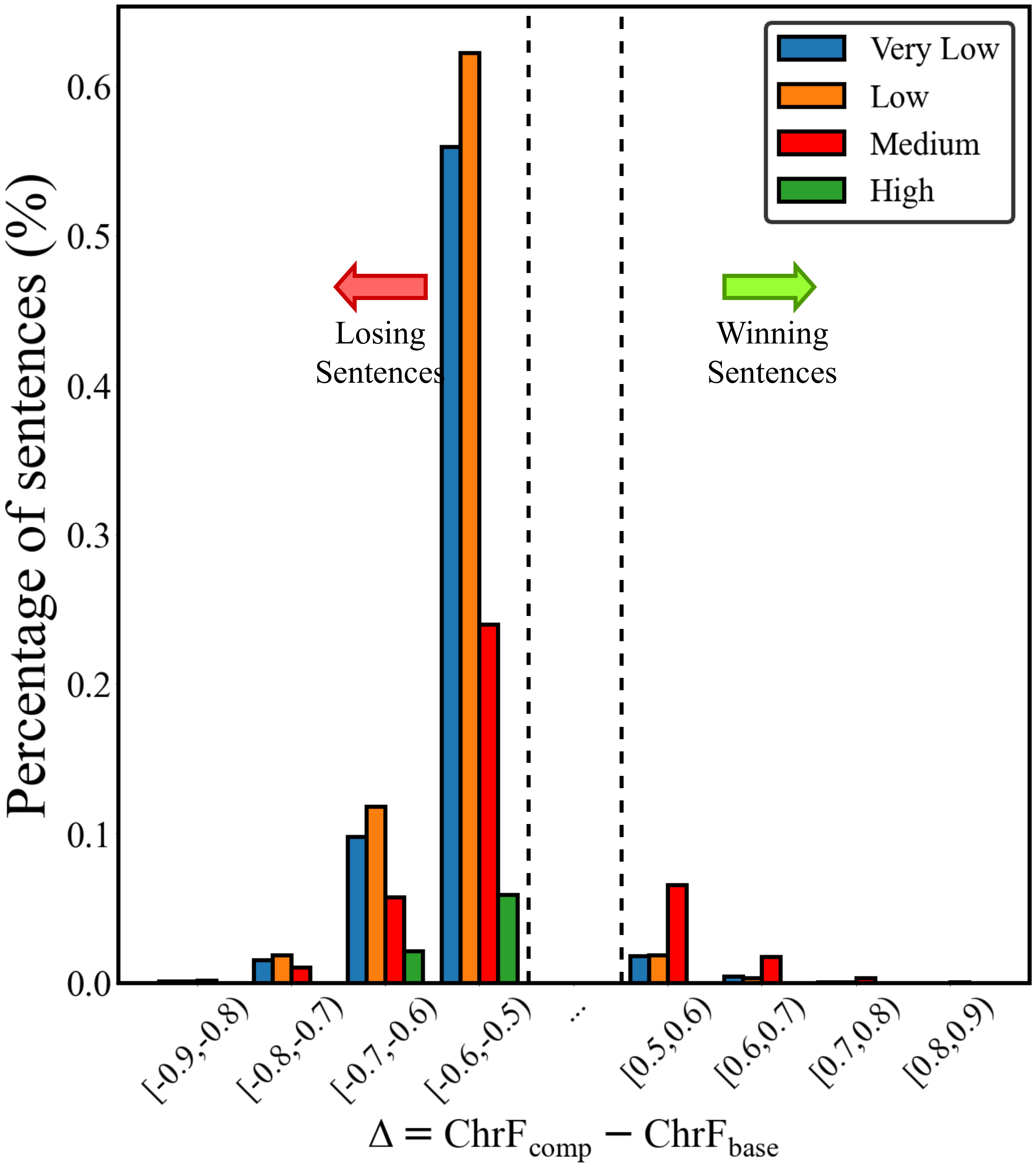}
\caption{Normalized distribution of sentences.}
\end{subfigure}\\[1ex]
\begin{subfigure}{\linewidth}
\centering
\includegraphics[width=\textwidth]{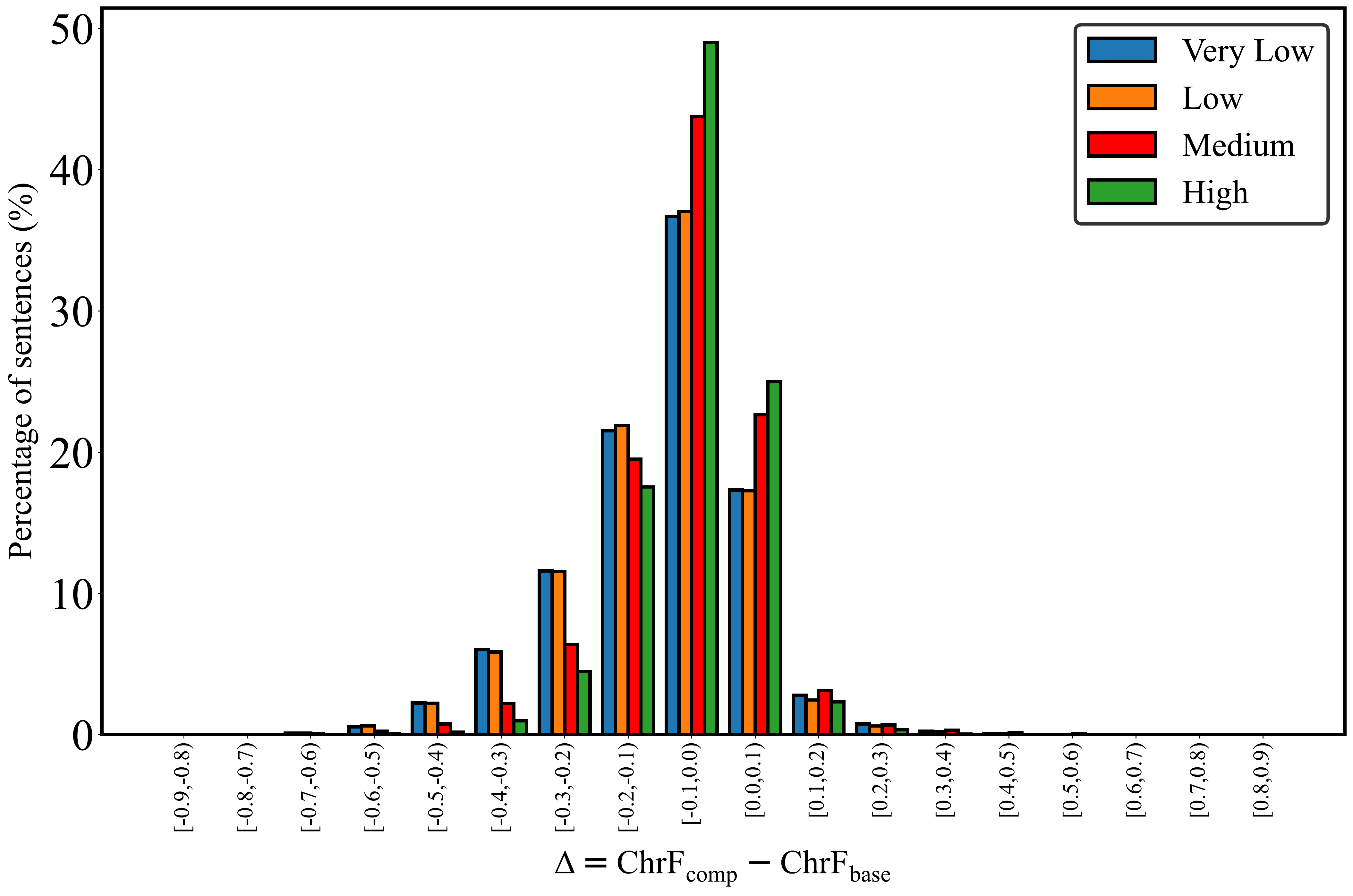}
\caption{Normalized distribution of sentences in each bin for different categories.}
\end{subfigure}
\caption{ChrF analysis of pruned 45$\%$ M2M-100 model.}
\end{figure}

\subsection{Quantized Model}
~~

\begin{figure}[!htb]
\begin{subfigure}{.5\linewidth}
\centering
\includegraphics[width=\textwidth]{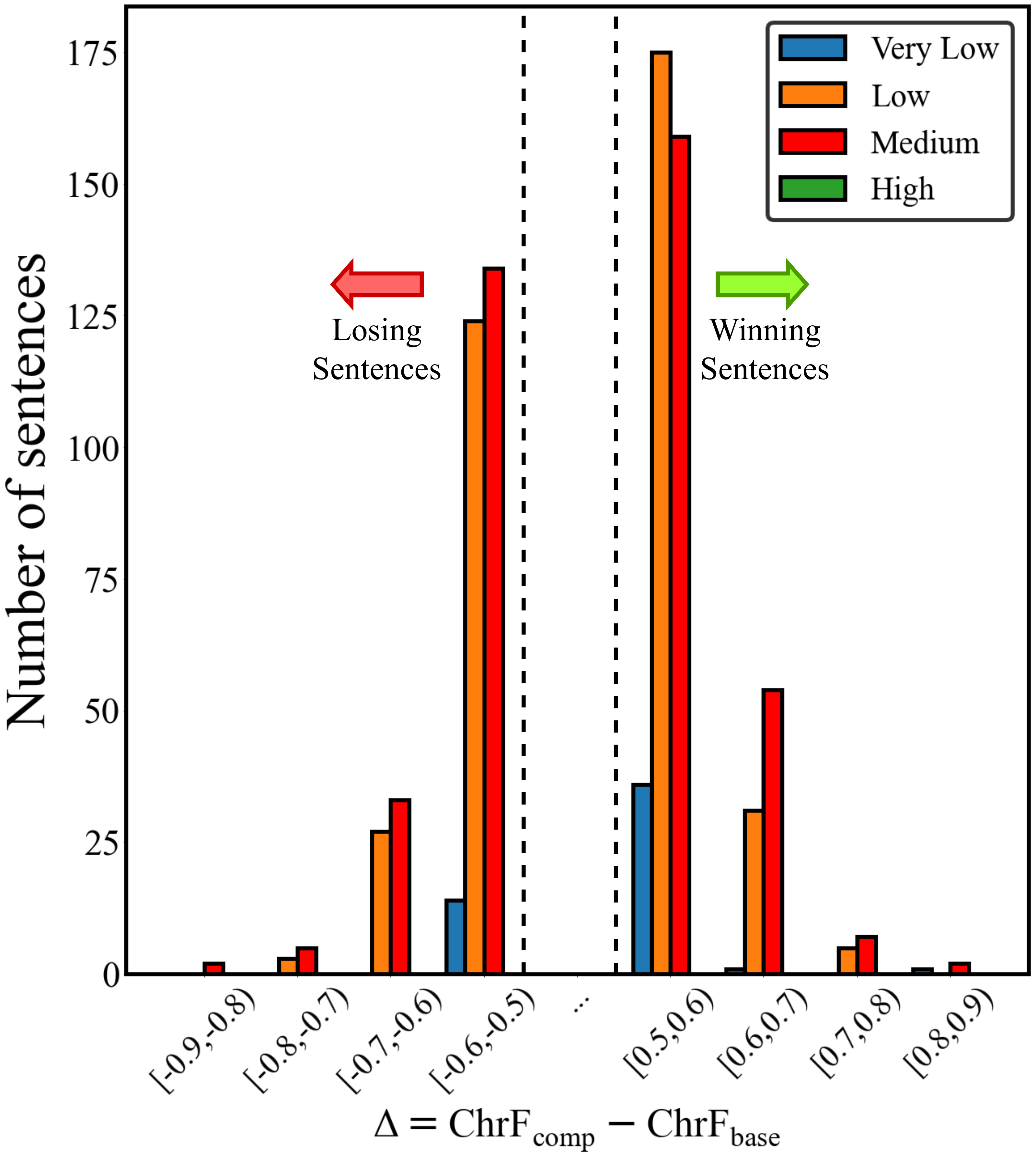}
\caption{Absolute number of sentences.}
\end{subfigure}%
\begin{subfigure}{.5\linewidth}
\centering
\includegraphics[width=\textwidth]{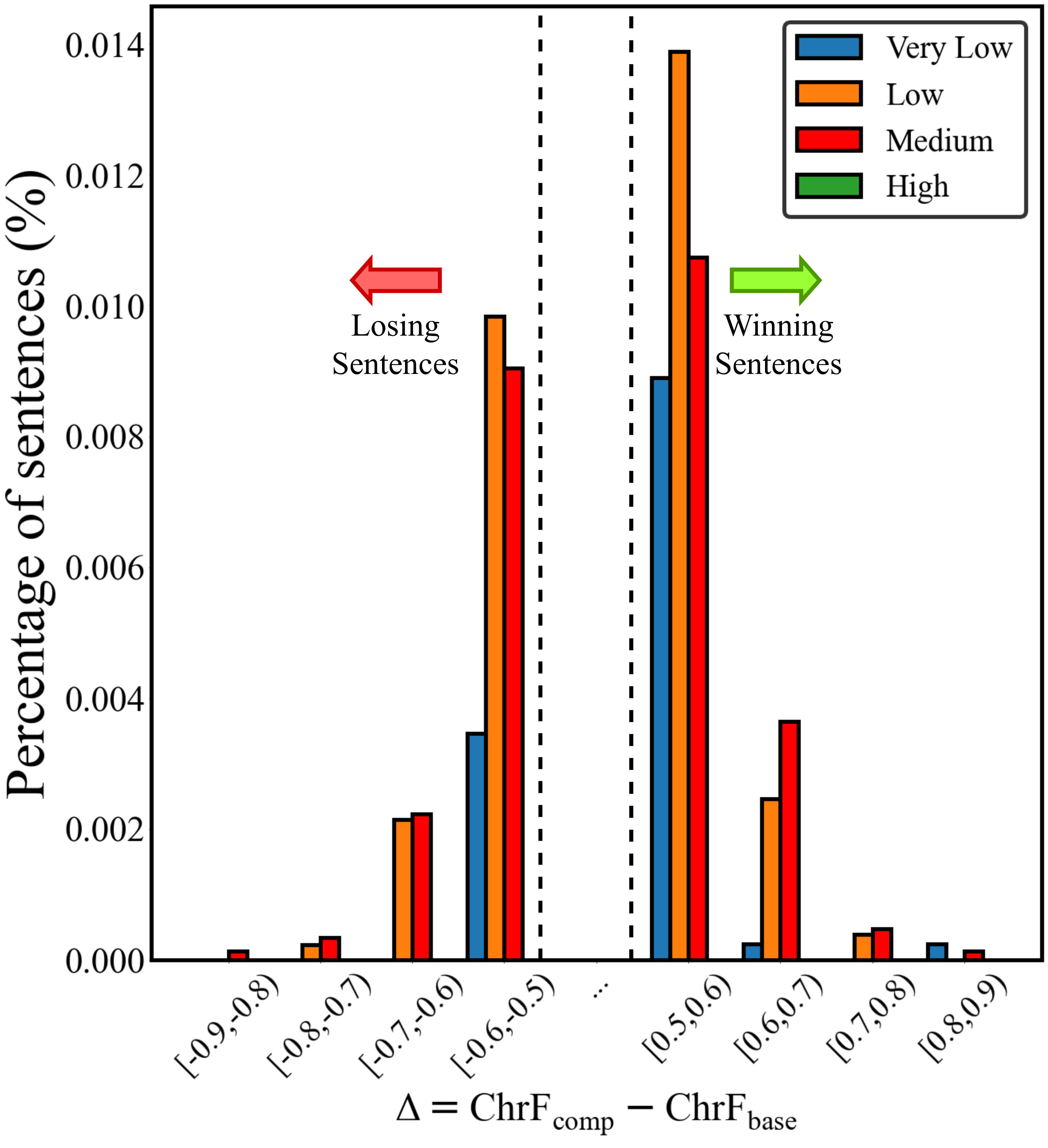}
\caption{Normalized distribution of sentences.}
\end{subfigure}\\[1ex]
\begin{subfigure}{\linewidth}
\centering
\includegraphics[width=\textwidth]{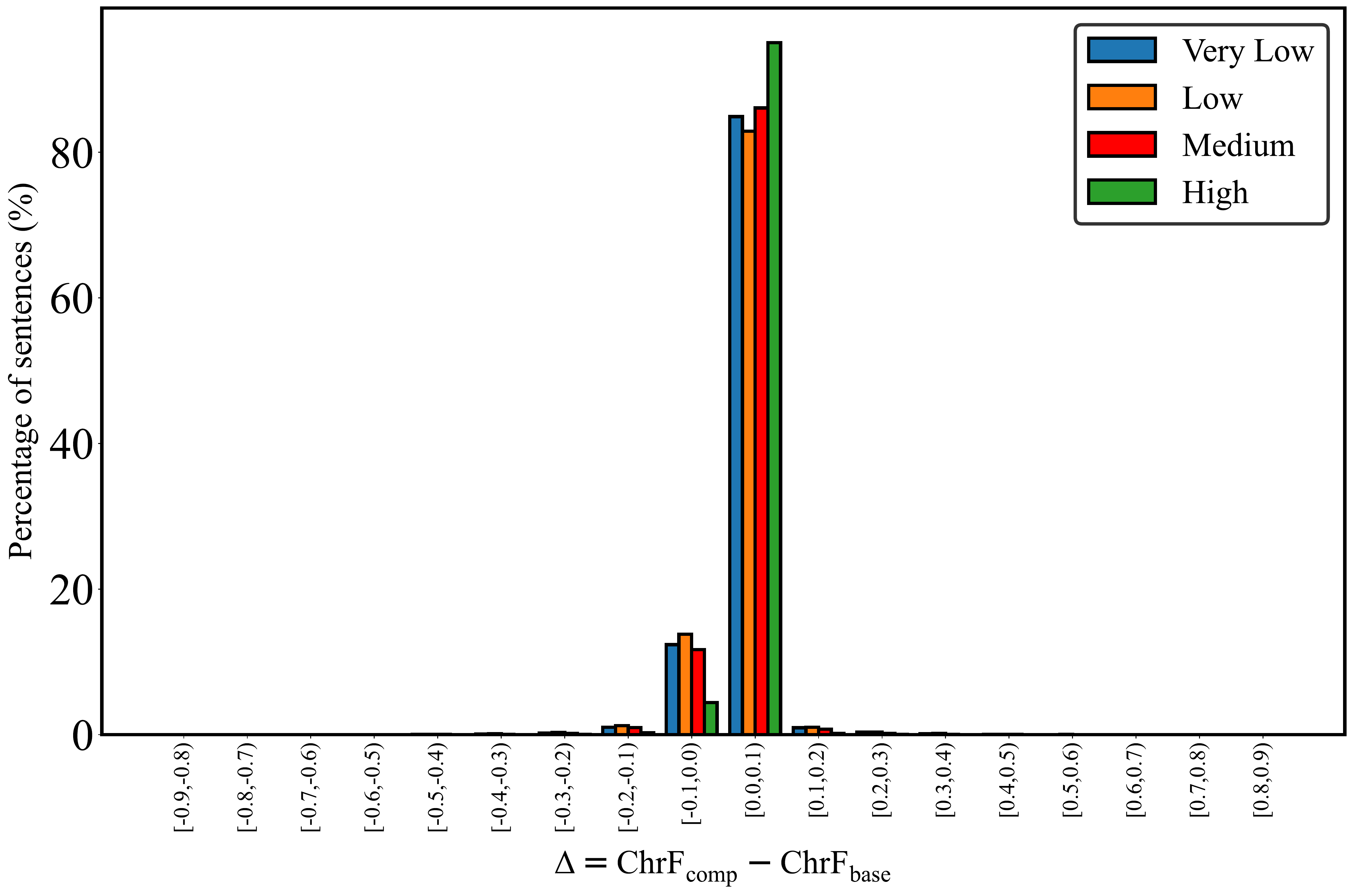}
\caption{Normalized distribution of sentences in each bin for different categories.}
\end{subfigure}
\caption{ChrF analysis of quantized M2M-100 model.}
\end{figure}

\section{Languages with Two Scripts in M2M-100 Training}
\label{app:flores-two-script}
~~
\begin{table}[htb]
\centering
\begin{adjustbox}{width=0.18\linewidth}
\begin{tabular}{l|l}
ISO & Language  \\ \hline
sr & Serbian  \\ \hline
cy & Welsh  \\ \hline
az & Azerbaijani \\ \hline
uz & Uzbek \\ \hline
ja & Japanese \\ \hline
bn & Bengali \\ \hline 
lo & Lao \\ \hline
zh & Chinese \\
\hline
\end{tabular}
\end{adjustbox}
\caption{Languages for which M2M-100 training data contains two scripts, while FLORES-101 provides one script for the evaluation.}
\end{table}

\section{Most Affected Language Pairs After Compression}
\label{app:most}

Language pairs are selected, if both quantization and pruning have significant effect on them~(based on spBLEU performance shown in Figure~\ref{fig:scatter}).

\begin{table}[htb]
    \begin{subtable}{.5\linewidth}
      \centering
      \begin{adjustbox}{width=0.4\linewidth}
        \begin{tabular}{l|l}
        Source & Target \\ \hline
        Catalan & Cebuano \\
        Latvian & Igbo \\
        Arabic & Igbo \\
        Danish & Xhosa \\
        French & Zulu \\
        \end{tabular}
        \end{adjustbox}
      \caption{Most losing language pairs}
    \end{subtable}%
    \begin{subtable}{.5\linewidth}
      \centering
      \begin{adjustbox}{width=0.5\linewidth}
        \begin{tabular}{l|l}
        Source & Target \\ \hline
        Latvian & Vietnamese \\
        Bulgarian & Latvian \\
        Arabic & Urdu \\
        Thai & Vietnamese \\
        Latvian & Italian \\
        \end{tabular}
        \end{adjustbox}
    \caption{Most winning language pairs}
    \end{subtable} 
    \caption{Most affected language pairs after the compression.}
    \label{app:tab:losing}
\end{table}

\section{Proposed Metrics for MT-Gender Benchmark}
\label{app:mt-gender:metric}

Equation~\ref{eq:mtgender1} considers the range of F1 scores for female and male subsets, while the simple difference between F1 scores does not reflect the range of F1 scores. The range is crucial since a model with the same F1 score difference but higher individual F1 scores should have a lower fairness score, as lied in Equation~\ref{eq:mtgender1}. \\
We also believe equation~\ref{eq:mtgender2} is a better metric than the simple difference between accuracies of the model in pro-stereotypical and anti-stereotypical subsets since it again considers the range of scores, and ignores missed translations and wrongly aligned genders. Additionally, it exactly reflects the difference in the behavior of the model in these two subsets. If the compressed model has a contrary performance in pro- and anti-stereotypical subsets, e.g. amplifying the bias in the anti-stereotypical subset more than the pro-stereotypical one or decreasing the bias more in one subset, then $\psi*$ becomes higher. We suggest using Equation~\ref{eq:mtgender1} and Equation~\ref{eq:mtgender2} for comparing models on MT-Gender benchmark~\cite{stanovsky-etal-2019-evaluating,kocmi-etal-2020-gender}.

\section{MT-Gender Results per Language}
\label{app:mt-gender}
~~
\begin{table}[!htb]
\begin{subtable}{0.33\textwidth}
\begin{adjustbox}{width=0.95\textwidth}
\centering
\begin{tabular}{l|c|c}
Model & $\psi$~($\%$) & $\psi^{\ast}$~($\%$) \\
\hline
Original M2M-100 & 21.01 & 15.09 \\
Pruned 30$\%$ M2M-100 & 20.71 & 16.87 \\
Pruned 45$\%$ M2M-100 & 28.58 & 17.33 \\
Quantized M2M-100 & 18.07 & 12.55 \\
\hline
\end{tabular}
\end{adjustbox}
\caption{Arabic}
\end{subtable}%
\begin{subtable}{0.33\textwidth}
\begin{adjustbox}{width=0.95\textwidth}
\centering
\begin{tabular}{l|c|c}
Model & $\psi$~($\%$) & $\psi^{\ast}$~($\%$) \\
\hline
Original M2M-100 & 39.02 & 11.39 \\
Pruned 30$\%$ M2M-100 & 45.19 & 7.15 \\
Pruned 45$\%$ M2M-100 & 45.56 & 18.54 \\
Quantized M2M-100 & 40.93 & 2.54 \\
\hline
\end{tabular}
\end{adjustbox}
\caption{Ukrainian}
\end{subtable}%
\begin{subtable}{0.33\textwidth}
\begin{adjustbox}{width=0.95\textwidth}
\centering
\begin{tabular}{l|c|c}
Model & $\psi$~($\%$) & $\psi^{\ast}$~($\%$) \\
\hline
Original M2M-100 & 7.98 & 20.09 \\
Pruned 30$\%$ M2M-100 & 10.38 & 16.30 \\
Pruned 45$\%$ M2M-100 & 8.89 & 2.75 \\
Quantized M2M-100 & 10.39 & 21.26 \\
\hline
\end{tabular}
\end{adjustbox}
\caption{Hebrew}
\end{subtable}%
\\
\par\bigskip
\begin{subtable}{0.33\textwidth}
\begin{adjustbox}{width=0.95\textwidth}
\centering
\begin{tabular}{l|c|c}
Model & $\psi$~($\%$) & $\psi^{\ast}$~($\%$) \\
\hline
Original M2M-100 & 29.06 & 3.93 \\
Pruned 30$\%$ M2M-100 & 29.10 & 2.30 \\
Pruned 45$\%$ M2M-100 & 30.28 & 8.08 \\
Quantized M2M-100 & 32.65 & 8.74 \\
\hline
\end{tabular}
\end{adjustbox}
\caption{Russian}
\end{subtable}%
\begin{subtable}{0.33\textwidth}
\begin{adjustbox}{width=0.95\textwidth}
\centering
\begin{tabular}{l|c|c}
Model & $\psi$~($\%$) & $\psi^{\ast}$~($\%$) \\
\hline
Original M2M-100 & 22.46 & 2.03 \\
Pruned 30$\%$ M2M-100 & 30.17 & 13.81 \\
Pruned 45$\%$ M2M-100 & 48.59 & 4.61 \\
Quantized M2M-100 & 24.71 & 2.6 \\
\hline
\end{tabular}
\end{adjustbox}
\caption{Italian}
\end{subtable}%
\begin{subtable}{0.33\textwidth}
\begin{adjustbox}{width=0.95\textwidth}
\centering
\begin{tabular}{l|c|c}
Model & $\psi$~($\%$) & $\psi^{\ast}$~($\%$) \\
\hline
Original M2M-100 & 13.86 & 28.71 \\
Pruned 30$\%$ M2M-100 & 29.03 & 40.20 \\
Pruned 45$\%$ M2M-100 & 38.44 & 32.83 \\
Quantized M2M-100 & 15.43 & 25.86 \\
\hline
\end{tabular}
\end{adjustbox}
\caption{French}
\end{subtable}%
\\
\par\bigskip
\begin{subtable}{0.33\textwidth}
\begin{adjustbox}{width=0.95\textwidth}
\centering
\begin{tabular}{l|c|c}
Model & $\psi$~($\%$) & $\psi^{\ast}$~($\%$) \\
\hline
Original M2M-100 & 5.77 & 15.72 \\
Pruned 30$\%$ M2M-100 & 4.89 & 14.62 \\
Pruned 45$\%$ M2M-100 & 22.53 & 34.01 \\
Quantized M2M-100 & 6.01 & 15.11 \\
\hline
\end{tabular}
\end{adjustbox}
\caption{Spanish}
\end{subtable}%
\begin{subtable}{0.33\textwidth}
\begin{adjustbox}{width=0.95\textwidth}
\centering
\begin{tabular}{l|c|c}
Model & $\psi$~($\%$) & $\psi^{\ast}$~($\%$) \\
\hline
Original M2M-100 & 6.48 & 16.93 \\
Pruned 30$\%$ M2M-100 & 13.16 & 26.83 \\
Pruned 45$\%$ M2M-100 & 22.14 & 18.12 \\
Quantized M2M-100 & 6.23 & 14.96 \\
\hline
\end{tabular}
\end{adjustbox}
\caption{German}
\end{subtable}%
\begin{subtable}{0.33\textwidth}
\begin{adjustbox}{width=0.95\textwidth}
\centering
\begin{tabular}{l|c|c}
Model & $\psi$~($\%$) & $\psi^{\ast}$~($\%$) \\
\hline
Original M2M-100 & 18.20 & 39.01 \\
Pruned 30$\%$ M2M-100 & 21.82 & 42.60 \\
Pruned 45$\%$ M2M-100 & 25.95 & 45.01 \\
Quantized M2M-100 & 18.24 & 38.42 \\
\hline
\end{tabular}
\end{adjustbox}
\caption{Polish}
\end{subtable}%
\\
\par\bigskip
\begin{subtable}{0.33\textwidth}
\begin{adjustbox}{width=0.95\textwidth}
\centering
\begin{tabular}{l|c|c}
Model & $\psi$~($\%$) & $\psi^{\ast}$~($\%$) \\
\hline
Original M2M-100 & 7.91 & 12.14 \\
Pruned 30$\%$ M2M-100 & 11.65 & 14.43 \\
Pruned 45$\%$ M2M-100 & 19.31 & 27.23 \\
Quantized M2M-100 & 9.78 & 13.26 \\
\hline
\end{tabular}
\end{adjustbox}
\caption{Czech}
\end{subtable}%
\caption{MT-Gender~\cite{stanovsky-etal-2019-evaluating,kocmi-etal-2020-gender} results for M2M-100 12B~\cite{m2m-100}, and compressed models.}
\end{table}

\newpage
\section{Detailed DiBiMT Results}
\label{app:dibimt}
~~
\begin{table}[!htb]
\begin{subtable}{0.5\textwidth}
\begin{adjustbox}{width=0.95\textwidth}
\centering
\begin{tabular}{l|c|c|c|c|c}
Model & SFII & SPDI & MFS & MFS$^{+}$ & Avg \\
\hline
Original M2M-100 & 89.14 & 80.59 & 41.8 & 92.59 & 76.03 \\
Pruned 30$\%$ M2M-100 & 87.32 & 80.56 & 39.55 & 93.04 & 75.11 \\
Pruned 45$\%$ M2M-100 & 86.78 & 82.9 & 39.93 & 92.41 & 75.50 \\
Quantized M2M-100 & 88.86 & 81.26 & 43.32 & 92.51 & 76.48 \\ 
\hline
\end{tabular}
\end{adjustbox}
\caption{Chinese}
\end{subtable}%
\begin{subtable}{0.5\textwidth}
\begin{adjustbox}{width=0.95\textwidth}
\centering
\begin{tabular}{l|c|c|c|c|c}
Model & SFII & SPDI & MFS & MFS$^{+}$ & Avg \\
\hline
Original M2M-100 & 80 & 71.61 & 60.63 & 89.76 & 75.5 \\
Pruned 30$\%$ M2M-100 & 78.96 & 73.79 & 61.44 & 88.56 & 75.68 \\
Pruned 45$\%$ M2M-100 & 81.28 & 77.05 & 62.5 & 91.67 & 78.12 \\
Quantized M2M-100 & 82.32 & 74.42 & 61.07 & 91.22 & 77.25 \\ 
\hline
\end{tabular}
\end{adjustbox}
\caption{German}
\end{subtable}%
\\
\par\bigskip
\begin{subtable}{0.5\textwidth}
\begin{adjustbox}{width=0.95\textwidth}
\centering
\begin{tabular}{l|c|c|c|c|c}
Model & SFII & SPDI & MFS & MFS$^{+}$ & Avg \\
\hline
Original M2M-100 & 75.99 & 70.53 & 61.23 & 88.41 & 74.04 \\
Pruned 30$\%$ M2M-100 & 75.91 & 71.86 & 60.92 & 87.74 & 74.10 \\
Pruned 45$\%$ M2M-100 & 83.38 & 75.08 & 62.22 & 86.67 & 76.83 \\
Quantized M2M-100 & 81.73 & 75.81 & 63.33 & 88.33 & 77.3 \\ 
\hline
\end{tabular}
\end{adjustbox}
\caption{Italian}
\end{subtable}%
\begin{subtable}{0.5\textwidth}
\begin{adjustbox}{width=0.95\textwidth}
\centering
\begin{tabular}{l|c|c|c|c|c}
Model & SFII & SPDI & MFS & MFS$^{+}$ & Avg \\
\hline
Original M2M-100 & 68.16 & 66.42 & 47.06 & 83.82 & 66.36 \\
Pruned 30$\%$ M2M-100 & 68.2 & 64.73 & 48.21 & 87.18 & 67.08 \\
Pruned 45$\%$ M2M-100 & 70.92 & 66.41 & 50 & 85.29 & 68.15 \\
Quantized M2M-100 & 68.16 & 69.03 & 44.19 & 86.51 & 66.97 \\ 
\hline
\end{tabular}
\end{adjustbox}
\caption{Russian}
\end{subtable}%
\\
\par\bigskip
\begin{subtable}{0.5\textwidth}
\begin{adjustbox}{width=0.95\textwidth}
\centering
\begin{tabular}{l|c|c|c|c|c}
Model & SFII & SPDI & MFS & MFS$^{+}$ & Avg \\
\hline
Original M2M-100 & 75.08 & 68.92 & 53.44 & 83.61 & 70.26 \\
Pruned 30$\%$ M2M-100 & 71.58 & 70.26 & 54.58 & 82.71 & 69.78 \\
Pruned 45$\%$ M2M-100 & 78.39 & 72.46 & 52.33 & 83.15 & 71.58 \\
Quantized M2M-100 & 76.45 & 69.72 & 56.88 & 85.63 & 72.17 \\ 
\hline
\end{tabular}
\end{adjustbox}
\caption{Spanish}
\end{subtable}%
\caption{DiBiMT~\cite{dibimt} evaluation for M2M-100 12B~\cite{m2m-100}, and compressed models.}
\end{table}

\end{appendices}

\end{document}